# Privacy-hardened and hallucination-resistant synthetic data generation with logic-solvers

Mark A. Burgess[1], Brendan Hosking[2], Roc Reguant[2], Anubhav Kaphle[3], Mitchell J. O'Brien[2], Letitia M.F. Sng[2], Yatish Jain[2,5], Denis C. Bauer[4,5,6]*

[1] Australian e-Health Research Centre, Commonwealth Scientific and Industrial Research Organisation, Canberra
[2] Australian e-Health Research Centre, Commonwealth Scientific and Industrial Research Organisation, Sydney
[3] Australian e-Health Research Centre, Commonwealth Scientific and Industrial Research Organisation, Melbourne
[4] Australian e-Health Research Centre, Commonwealth Scientific and Industrial Research Organisation, Adelaide
[5] Macquarie University, Applied BioSciences, Faculty of Science and Engineering, Macquarie Park, Australia
[6] Macquarie University, Department of Biomedical Sciences, Macquarie Park, Australia

**Abstract:**

Machine-generated data is a valuable resource for training Artificial Intelligence algorithms, evaluating rare workflows, and sharing data under stricter data legislations. The challenge is to generate data that is accurate and private. Current statistical and deep learning methods struggle with large data volumes, are prone to hallucinating scenarios incompatible with reality, and seldom quantify privacy meaningfully. Here we introduce Genomator, a logic solving approach (SAT solving), which efficiently produces private and realistic representations of the original data. We demonstrate the method on genomic data, which arguably is the most complex and private information. Synthetic genomes hold great potential for balancing underrepresented populations in medical research and advancing global data exchange. We benchmark Genomator against state-of-the-art methodologies (Markov generation, Restricted Boltzmann Machine, Generative Adversarial Network and Conditional Restricted Boltzmann Machines), demonstrating an 84-93% accuracy improvement and 95-98% higher privacy. Genomator is also 1000-1600 times more efficient, making it the only tested method that scales to whole genomes. We show the universal trade-off between privacy and accuracy, and use Genomator's tuning capability to cater to all applications along the spectrum, from provable private representations of sensitive cohorts, to datasets with indistinguishable pharmacogenomic profiles. Demonstrating the production-scale generation of tuneable synthetic data can increase trust and pave the way into the clinic.

## Introduction

Synthetic data is emerging as a valuable resource in the medical sector [1], driven by stricter data sharing and privacy regulations[2], as well as the need for bioinformatics standardisation through characterized reference samples[3]. Synthetic genomes are of particular interest due to the high cost of obtaining data through sequencing and the risk of exposing personally identifiable or sensitive information when sharing real genomic data. Use cases for synthetic genomes include (a) "digital twin" dataset[4] where population-specific genomic information is reproduced without replicating any one individual specifically[5], (b) "Truth Challenges", where synthetic genomes are spiked with difficult to identify variants for grading pathology providers[6], and (c) methodology benchmark for evaluation[7], or for capturing the "long tail" of human diversity and disease states[1].

Gartner expects synthetic data to "completely overshadow real data in AI models by 2030"[8], illustrating the need for creating high quality synthetic data to ensure the subsequently trained models have fidelity, generalisability, diversity, and compliance. We discuss three key considerations for this. *Firstly*, the risk of producing skewed, incomplete, or discriminatory synthetic genomes exists if the generative algorithms fail to adequately reflect the diversity and complexity of features (or patterns) of the real data[9]. *Secondly*, privacy is a multifaceted problem, where exposing even a single nucleotide can reveal sensitive information, e.g. breast cancer risk[10], or expose the identity of an individual through a rare private variant[11]. Explainability of generative algorithms is hence desirable for accountability and building trust. *Lastly*, resource consumption is a critical element that can severely limit application scope and sustainability[12].

Traditional approaches for creating synthetic genomic information (such as HapGene2[13] and Hapnest[14]) generate novel haplotypes by resampling from reference genomes, based on a stochastic model that simulates the underlying processes of coalescence, recombination, and mutation. These methods require additional information, for example genetic parameters related to evolutionary history such as recombination rate, allelic age, mutation rate and can hence be biased if the wrong assumptions were made. Furthermore, this information is typically not available for large genomic cohorts.

More recent methods developed by Yelmen *et al.* [15] are unsupervised, include approaches like Markov Models, Generative Adversarial Networks (GANs) and Restricted Boltzmann Machines (RBMs) [15]. Their most recent paper[16] concludes that GANs provide the most faithful representation of real data out of their tested methods. However, there are well documented problems in the GAN training process, such as mode collapse[17] resulting in a reduced output range because the model has limited its training to subsets of the input data, as well as the need for large volumes of data and compute resources to train on high dimensional genomic data [18]. Finally, privacy is currently not included as part of the learning process, leading to potential information leakage.

In contrast to resampling and learning approaches, SAT solvers are a class of algorithms to solve mathematical and combinatorial problems that involve the (SAT)isfaction of constraints among Boolean variables[19]. We show how SAT solvers can be used to deductively generate synthetic data and address the resource and privacy issues of the previously published methods. Notably, SAT solvers have been shown to be NP-complete[20] and can rapidly solve (in polynomial time) problems where all constraints can be represented as between two Boolean variables (2SAT). While the genomics problem is not exclusively 2SAT by including a small number of size-4 clauses, we have shown that it is practical to apply it to whole genomes. Although SAT solvers have been variously applied to challenges in biology, such as haplotype inference[21,22], optimisation of breeding schemes[23], computing genomic distances[24], and gene regulatory network inferences[25], they are regarded as being underutilised in computational biology[26].

In this manuscript we present Genomator, a resource efficient SAT solving application for *the de-novo* creation of synthetic data. Genomator is hallucination-resistant and able to tailor accuracy vs privacy qualities to the application at hand. We systematically compared and evaluated the accuracy, privacy, and utility of Genomator against state-of-art approaches: Markov chain generation, Generative Adversarial Networks (GAN), Restricted Boltzmann Machines (RBM) and Conditional Restricted Boltzmann Machines (CRBM).

## Methods

### 2.1 Datasets

For evaluating population-structure we used the 805 SNPs from across the genome as selected by Yelmen *et al.* [15] to represent the population structure of the 2504 samples in the 1000 Genomes

Project (1KGP)[27]. Principal component analyses (PCA) were conducted using the SciKit-Learn PCA algorithm and plotted using MatplotLib python libraries. Please note, we follow Yelmen *et al.'s* approach of only selecting one of the two chromosomes for the 805 SNP data. For all other datasets we use phased information.

For evaluating linkage disequilibrium (LD) we used the latest release of the 1KGP and selected the larger genes that all methods could process – RNA binding fox-1 homolog 1 (*RBFOX1*), Fragile Histidine Triad Diadenosine Triphosphatase (*FHIT*), AGBL carboxypeptidase 4 (*AGBL4*) and coiled-coil serine rich protein 1 (*CCSER1*) genes, see **Supplemental Table 4**. For each gene, including introns, exons, and untranslated regions, a pre-processing step was performed on genetic variants from the latest version of the 1KGP project containing 3201 samples using PLINK2[28]. The process included the retention of variants with a Minor Allele Frequency (MAF) greater than or equal to 0.01, the exclusion of variants showing deviations from Hardy-Weinberg Equilibrium (HWE) with a significance threshold set at 1e-06, and the removal of duplicate variants from the dataset.

For evaluating the scalability of different methods, we generated data over increasing sections of the full 11 million SNPs (excluding sex chromosomes), using the same filtering, from the latest release of the 1KGP project using the first 400 samples of the 3201.

For calculations of pharmacogenetic SNP allele frequencies, chromosomes 10 and 16 were selected as they housed three important pharmacogenetic SNPs: *rs*4244285, *rs*4986893, and *rs*9923231 without Hardy-Weinberg Equilibrium (HWE) filtering. The PCA was conducted on the latest release 1KGP samples using PLINK2 [28] and Genomator generated 1000 samples projected onto the calculated PC space and ethnicity determined based on clustering with the 1KGP annotated 'Super Populations'.

## 2.2 Software tools

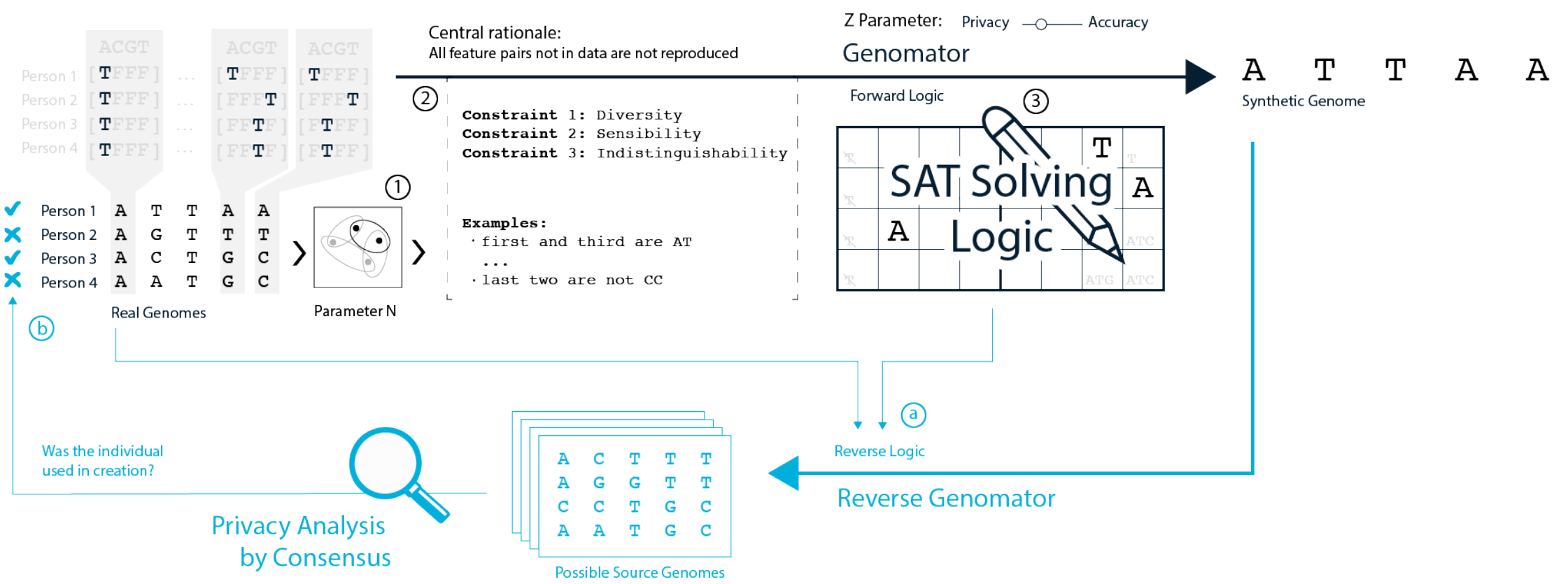

Figure 1: An illustration of Genomator (black) and Reverse Genomator (blue). From a cohort of real genomes, we first create overlapping clusters of size N (1). From a randomly chosen cluster, constraints are formed based on observations in the data (2) as illustrated with Booleans in the grey boxes and the examples about feature pairs. These constraints are fed into the SAT Solver (3) with the Z parameter controlling the accuracy-privacy trade-off. The sudoku-puzzle is an analogy for the logic solving approach. From this a synthetic genome is created and the process can be repeated to create more genomes from different clusters covering the diversity across the real genomes. Reverse Genomator (a) can take the synthetic genome, and together with the original real data and the constraints used, determine a set of possible source genomes from which privacy is assessed, such as which individual or set of individuals were used to create the synthetic genome (b). For illustration purposes we show single nucleotides, but the algorithm can process diploid genomes. For a more detailed illustration see **Supplemental Figure 1 and 4**.

**Genomator** constructs and solves a SAT problem to produce synthetic genomic data using real genotypic information captured in a Variant Call Format (VCF) file. The central rationale of Genomator's algorithm is to ensure that pairs of features that do not occur in the input data will also not occur in the synthetic data (see **Figure 1**).

The task is performed by constructing a set of constraints from the real genomes for the SAT solver to resolve into synthetic genomes. Genomator hence does not require extensive training but instead creates synthetic genomes iteratively from subsets of input data. This subset is picked at random from a set of clusters of size N created based on Hamming distance between the pairwise genotype vectors. This process can be repeated and is efficient enough to create tens of thousands of synthetic genomes, with the randomness in the clustering process creating overlapping clusters, which ensures that no sample is under-represented and diverse population-scale cohorts can be created.

There are two parameters, N and Z, controlling the size of the input clusters, and the strength of the attenuation of rare feature pairs, respectively. Increasing N makes the formed constraints less stringent as there are likely less feature pairs that none of the input data possess. The parameter Z can be randomised with each feature pair considered. Increasing Z hence stochastically reduces the contribution of rare (and potentially identifying) combinations of SNPs in the constraint creation and thus enhances privacy.

**Supplemental Section 1** explains the workflow in full, including pseudocode and detailed diagrams. Genomator is implemented using PySAT library[31] and, unless otherwise stated, is run with a cluster size of N = 10 and Z = 0.

**Reverse Genomator** complements Genomator and deductively determines possible combinations of input genomes (of size N) which could have been used by Genomator to construct the produced synthetic data (**Figure 1**). Reverse Genomator reverses Genomator's logic by constructing a SAT problem with constraints that eliminate input combinations from which Genomator could not have generated the synthetic data with.

**Markov Chain** method as implemented by Yelmen *et al.*[15] was used with a window size parameter of 10 variants.

**Restricted Boltzmann machine** (RBM) as implemented by Yelmen *et al.*[16] was used with a hidden layer size of 500, 50 Gibbs steps and 500 parallel chains to compute the negative term of the gradient in the training process, with a minibatch size of 500 and a learning rate of 0.01, data was trained till 1200 epochs.

**Conditional Restricted Boltzmann machine** (CRBM) implementation from Yelmen *et al.*[16] was used with a hidden layer size of 500, 50 Gibbs steps and 500 parallel chains to compute the negative term of the gradient in the training process, with a minibatch size of 500 and a learning rate of 0.005, data was trained till 1200 epochs with a window overlap of 300 (reduced to from the originally published value of 5000 to fit our dataset).

**Generative adversarial network** (GAN) implementation by Yelmen *et al.* from 2021[15] where they used a latent size of 600, generator learning rate of 0.0001, discriminator learning rate of 0.0008, with 2 intermediate layer dense neural networks for generator and discriminator with sizes a factor 1.2, 1.1 and 2, 3 smaller than the data's dimension respectively, with leakyReLU neural activation function with an alpha value of 0.01, GANs were trained for the suggested 20000 epochs - for investigation of the variability in this method we also optionally added a multiplier on the size of all neural layers. Please note that the newer Wasserstein GAN (WGAN) by Yelmen *et al.*[16] had an architecture tailored to the input SNP size and could not be executed on our data. We hence used GANs as a proxy.

For further details on all methods see **Supplemental Section 3**.

**2.3 Accuracy calculation**

**Wasserstein distance is used** for analysing the consistency between datasets. We employ the Sliced Wasserstein distance algorithm in the Python Optimal Transport library[30]. In Section 3.1, to quantify the **differences in the PCA** of the synthetic versus real data on the Yelmen 805 SNP dataset, we calculated the Wasserstein distance on the first two principal components between synthetic and real. In Section 3.4 we **evaluate the data itself**, hence calculated the Wasserstein distance across all dimensions between input and synthetic data and reported percentage error as proportion of the maximal of 800 haplotypes (400 SNPs of ploidy 2).

**LD correlation error** used for Figure 2b was calculated from Pairwise LD and evaluated using the Rogers-Huff r-squared method[29], which is part of the Scikit-Allel Python package. LD reproducibility between a synthetic and real dataset is quantified by average LD square error, obtained by subtracting the pairwise synthetic LD values from the true LD relationship and squaring the result and averaging. We evaluated synthetic data LD reproducibility generated from a training set against a test set - produced by equal split partition of datasets. We also calculated the difference in LD between synthetic and real data for all SNP pairs in a window size averaged across the tested genome. We systematically test window size ranges from 5000 to the maximal gene length in 5000 increments.

**2.4 Privacy calculation**

**Attribute inference-based privacy evaluation** emulates an inference attack scenario. We split the original genome dataset into two random subsets and generate synthetic data from each. Next, we compute the median Hamming distance between the input sequences and to the synthetic dataset created from the subset containing the input sequence (in-data) and the subset that does not contain it (out-data). The difference between the in-data and out-data distances reflects the additional likelihood that an attacker correctly identifies SNP attributes. This approach reflects Giomi *et al* [32] (inference attack) except baselining performance against synthetic data from a hold-out set, instead of random imputations.

**Private and fictitious quadruplet revelation** measures privacy by how likely rare combinations of SNPs are reproduced in the output. To do this, we randomly select SNP combinations in groups of four (quadruplets) and assess if they are private – meaning the combination occurs at most once in the original data - or fictitious, meaning it does not appear at all in the original data. We continue sampling until we have 100K fictitious and private combinations each. Next, we generate 1K versions of synthetic data and determine how many of the 100K SNP combinations occur across these 1K datasets to compute the likelihood for each category, i.e., private or fictitious. This reflects Giomi *et al* [32] ('singling out') except identifying unique combinations among the input data for comparison to the synthetic data, instead of *vice versa*.

**Exposure risk** deduces privacy directly and is unique to our logic-based approach. We randomly select *G* SNPs and *N* individuals from a real dataset (1KG project containing 3.2K individuals). We then generate 500 distinct plausible input combinations (of same size *N*) for a synthetic genome created from the same real cohort using Reverse Genomator. We then identify individuals present in all 500 combinations and label them as *exposed*. This process is repeated across multiple iterations, and the probability of exposure is modelled as a binomial event. The exposure risk is defined as the proportion of iterations where an individual appears in all reconstructed combinations, with confidence intervals calculated based on the binomial model. Please note, in a hypothetical scenario where an attacker is looking for “likely enough” cases, a Bayesian analysis can be performed as given in Supplemental Theorem 1.

## Results

### 3.1 Accuracy: SAT-solvers replicate both genome-wide and local genomic structure.

First, we evaluate the ability of the methods to produce synthetic data that captures the higher-order complexities of genomic data. We ran a principal component analysis (PCA) on the synthetic data generated from the Yelmen 805 SNP dataset (see **Section 2.1**) to measure how well each reproduces the well-known “V” shape of the underlying population structure. At this time synthetic samples were generated on a training set and plotted against a holdout set of individuals from the same cohort. **Figure 2a** shows that Genomator reproduces the population structure accurately (average Wasserstein Score of 1382±326) whereas RBM (1517±624), GAN (2010±497) produce synthetic data that are dispersed and slightly shifted from the real data**.** Markov Chain (7249±53) and CRBM (9751±109) perform poorly, showing clustering of synthetic samples and loss of the characteristic V shape. **Supplemental Tables 1a-e** illustrates the stability of the methods over 10 runs, the PCA of the median run is plotted in **Figure 2a** and PCA on larger datasets are in **Supplemental Section 11**.

We evaluate the methods’ abilities to replicate local interactions as these are of particular importance to medical and research applications[33]. We produced 1000 synthetic genomes for each method, focussing on the largest genes reproducible by all models (*RBFOX1*, *FHIT*, *AGBL4* and CCSER1), and compared the linkage disequilibrium (LD) patterns constructed to the original dataset. **Figure 2b** shows that for *AGBL4 gene*, Genomator captures both short distance LD as well as long distance LD. Markov chains only reproduce short distance LD, while RBM, CRBM and GAN capture both short- and long-range interactions but with visibly lower accuracy than Genomator. We quantify this for all genes in **Figure 2c** by calculating the average square error between the real and reproduced LD with increasing window sizes across the gene, i.e. small windows evaluate the capability to capture local SNP-SNP effects, while larger window sizes cover local as well as long-ranged interactions. Genomator has an average square error percentage of 2% (std 0.13%) over all genes compared to RBM of 11.4% (3.6%), CRBM of 17.1% (9.4%), Markov Chain of 21.6% (15.3%), and GAN of 19% (14%); for a detailed breakdown see **Supplemental Table 2a and b** with correlation measures provided in **Supplemental Table 3**.

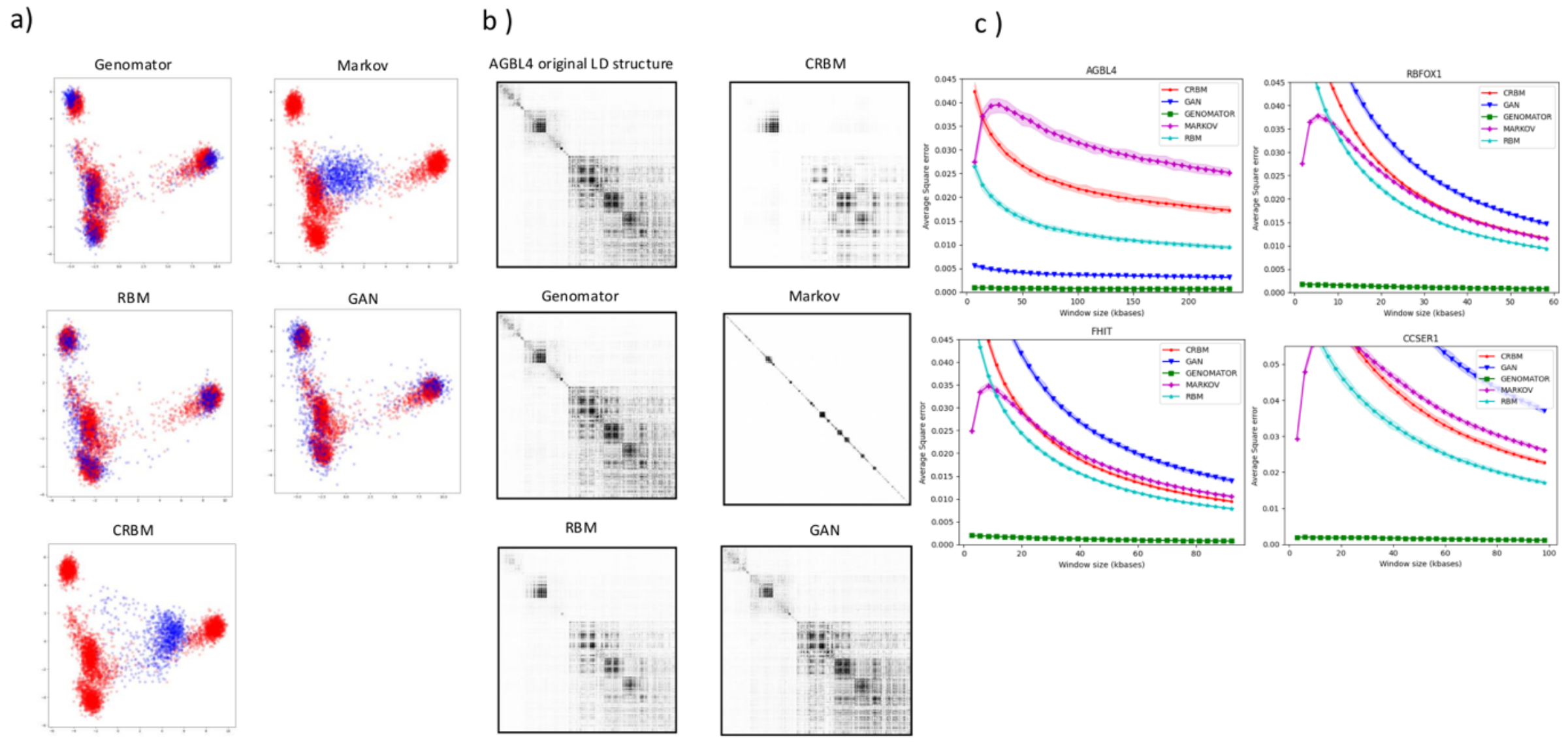

Figure 2: a) PCA of the first two principle components of the 805 SNP dataset (shown in Red) compared to the synthetic genome data generated by the 5 different methods (shown in Blue), b) LD of *AGBL4* (first 2000 SNPs*)* between loci in the original dataset and the synthetic data c) Average Square Error between observed and synthetic LD across the lengths of the gene for the four tested large genes across methods, where 95% confidence intervals were approximated using bootstrap resampling.

### 3.2 Efficiency: Runtime Statistics and trends

We evaluate the runtime of generating a synthetic genome from each of the methods. We included all training and processing, but excluded input and output file operations, as these are dominated by the initial loading of the VCF file and final writing of the synthetic data, which is common to all methods and increases with file size. We report the runtime for creating a synthetic genome from increasing input file sizes, up to the full 11 million SNPs, constituent of the full human genome (excluding sex chromosomes) from the 1KGP project. For this experiment we produced different VCF files of 400 human individuals with increasing genome segments across chromosomes 1-22 with MAF > 0.01 filtering. We ran all tests with Intel Xeon(R) CPU Platinum 8452Y, 2.00GHz, 36 cores, 80 GB of ram, with GPU Nvidia H100 94GB graphics card and two days of compute time.

**Figure 3** shows that Genomator is the only method scalable to the largest file size – producing synthetic genome across all human chromosomes in 18 seconds. RBM&CRBM ran out of graphics card VRAM above 1.6M SNPs, GAN ran out of VRAM above 6.4K SNPs, and Markov method ran out of RAM above 6.5M SNPs. On the 1638K SNPs completed by all but GANs, Genomator is 1067 to 1693 times (5 seconds vs 5283 – 8383 seconds) faster than the other tested methods – **see Supplemental Table 5**.

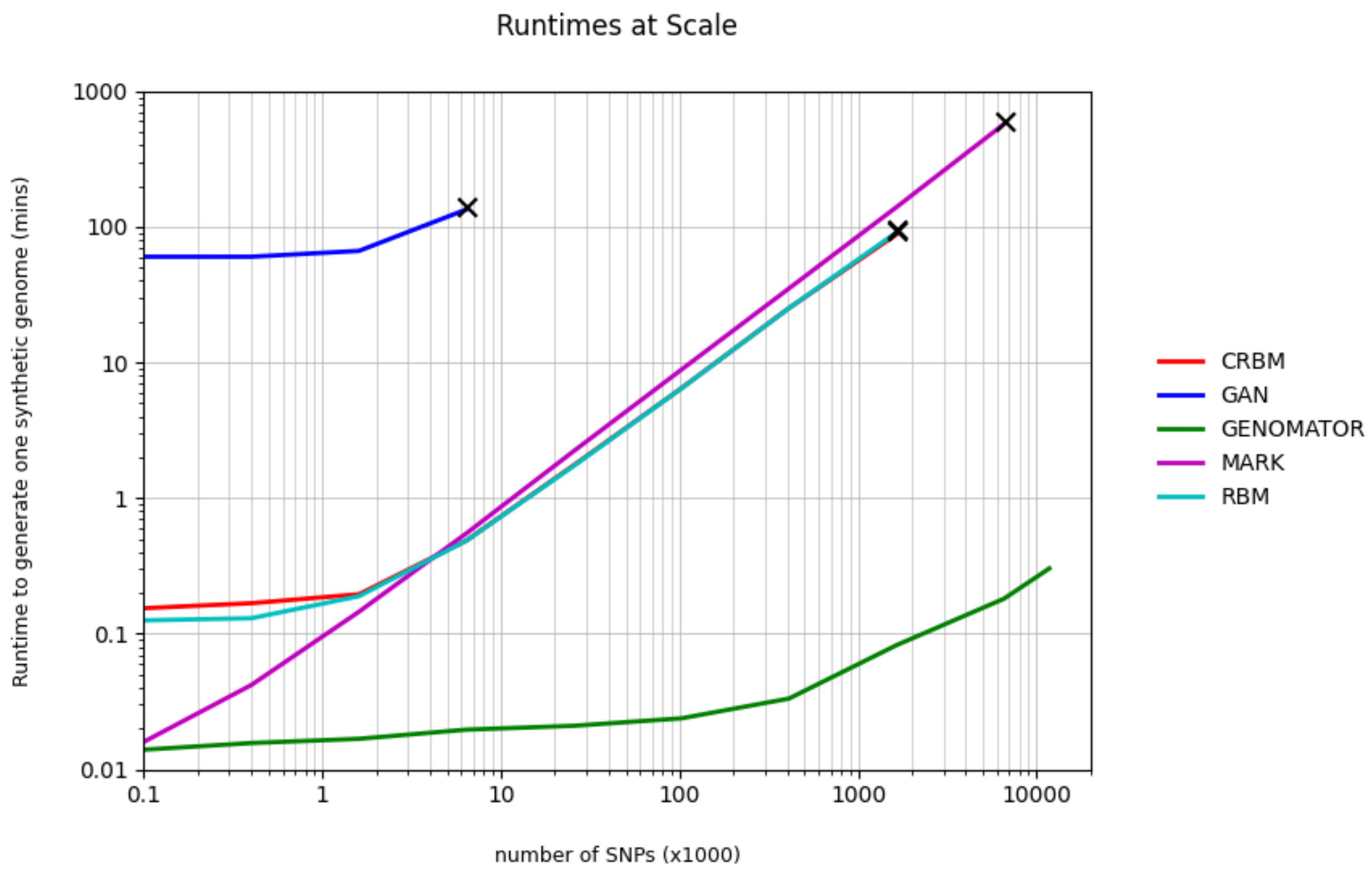

Figure 3: Runtime to create data of increasing size up to the full human genome of 11 million SNPs (excluding sex chromosomes). Failure point from consuming all available RAM and/or VRAM for the various methods is represented by crosses, noting that Genomator did not experience failure point, and executes the full 11M SNPs while only consuming 14 GB of the 80 GB allotted RAM.

### 3.3. Privacy

Many approaches use similarity tests, including Yelmen *et al.* [16] who uses nearest neighbour adversarial accuracy for privacy evaluation. This is based on the balanced accuracy of an adversarial classifier that attempts to determine if a given data point is *closer* to the real data cluster than to the synthetic cluster[34]. However, Stadler *et al.*[35] showed that these methods underestimate the privacy risks. We hence considered three approaches (based on attribute inference, private quadruplet revelation, and a direct deductive privacy inspection of the output) as introduced over the next sections.

Firstly, we simulate an attribute inference attack, where an attacker knows a subset of a target's genome and uses the nearest neighbour in the synthetic cohort to infer the remaining SNPs, similar to Giomi *et al* [32]. We calculated how accurate such an attacker would be on the datasets generated over the Yelmen 805 SNP dataset, as the down-scaled representation of the genome.

We split the data into two similar subsets by random selection. We calculated the "in-data" distance between the real data and the synthetic data generated from the subset containing the target individual, while the "out-data" distance was calculated on the cohort without the target individual using Hamming distance (**Supplemental Figure 9**). Ideally, in-data distance is small indicating a good fit, while the difference between in-data and out-data distance is zero. The difference between in-data and out-data distances can be interpreted as a measure of privacy, as the additional likelihood that an attacker could make a correct inference about a SNP of a target sequence owing due to the synthetic data being generated from a dataset including that target sequence.

**Figure 4** reports the performance for each method where we generate 2504 genomes using a range of parameter settings of the different Method (e.g. CRBM hidden-layer 100-2000, Markov window-size 2-22, GAN neural layer size multiplier 0.1-4.6, RBM hidden layer size 100-2000 and learning rate 0.01-0.005, Genomator N 15-280 and Z 1-54) see **Supplemental Table 8**. In the figure, each method creates a unique "frontier" as there is a trade-off between accuracy and privacy leakage. While Genomator and Markov's parameter setting allow the "frontier" to be traced out, the other methods are more static in their capacity to fit to the input data creating less flexibility in the outcome. The theoretical ideal is 0,0 and the minimum distance on this graph to this ideal point is calculated for each method - as shown in **Supplemental Table 9** and between these distances other methods are 22.9-73.8% further from the ideal point.

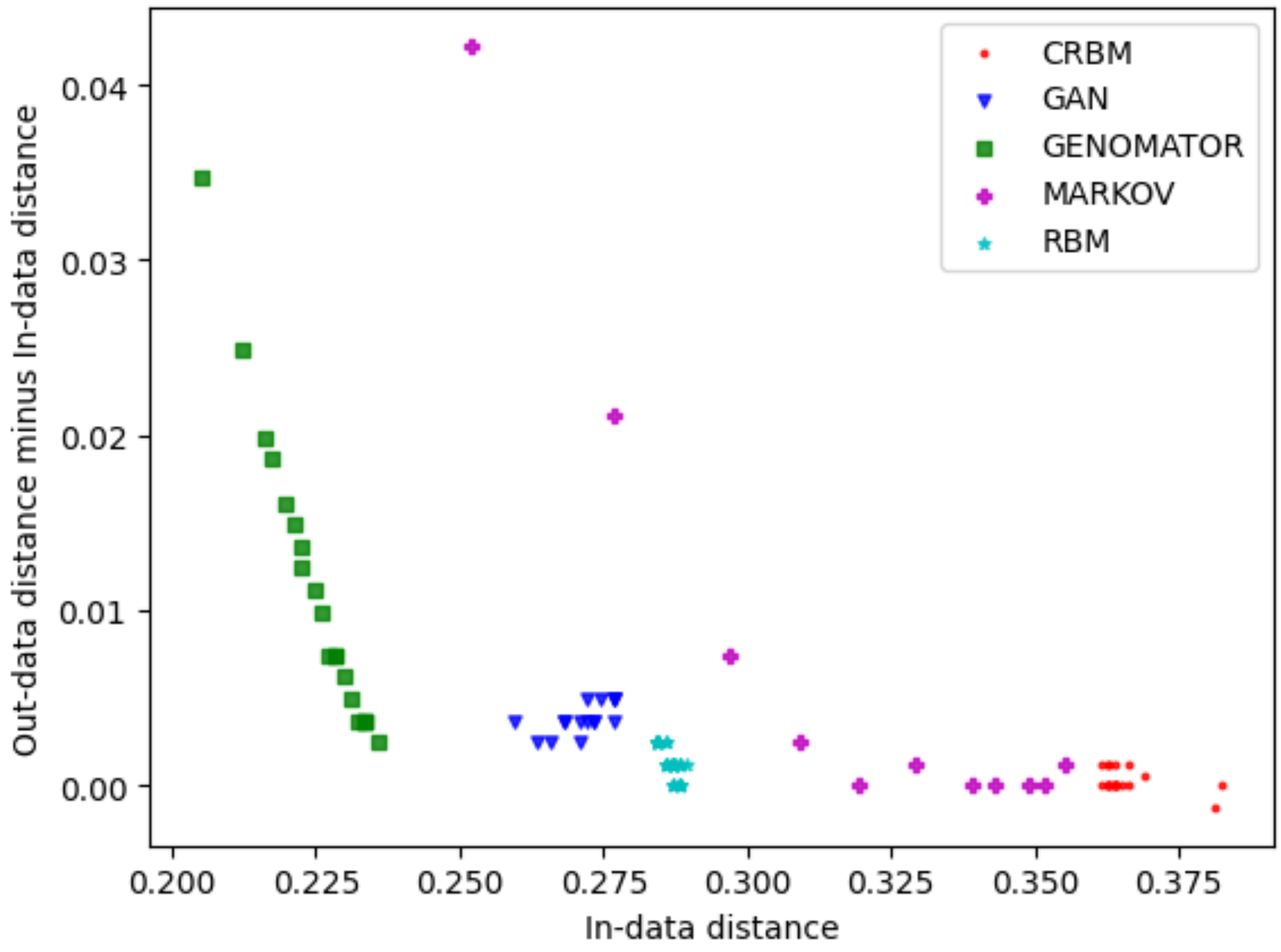

Figure 4: Results of SNP inference experiment on 805 SNP data for different methods. The ideal outcome is a perfect fit (in-data = 0) and no difference between in- and out-data distance.

### 3.4 Genomator can customise the level for accuracy-privacy trade-off

Currently, privacy is estimated through proxy-based methods as it is impractical for iterated learning methods like GANs and RBMs to deductively infer whose data was used in the synthetic data generation process. Since Genomator is a logic-based approach, reverse logic is possible. We hence developed Reverse Genomator, which identifies for any synthetic genome, the logical space of all possible combinations of subsets of input data that could have been used to generate it (given the full information about Genomator and a dataset from which Genomator's cluster input is selected from). Individuals who appear in all these subsets are deduced to have been used as input and are considered 'privacy exposed' (see Methods 2.4).

As noted in the previous section, there is a natural trade-off between accuracy, i.e. copying the data, and privacy. The new ability to quantify absolute privacy in addition to accuracy enables the tailoring of the accuracy-privacy ratio to the application at hand. To do this, we introduce the parameter Z, which defines the degree of randomised strengthening in Genomator's constraints, see **Method Section 2.2**.

To demonstrate the approach, we randomly select 400 SNPs from 400 distinct samples from *AGBL4* gene (3202 samples of 1KGP). We run Genomator to create synthetic data and then use Reverse Genomator to randomly reconstruct 500 distinct plausible input sets (of size 400 samples) that could have been used to create the synthetic data. If an individual is part of all reconstructions, the person is at risk of being 'privacy exposed'. To evaluate the widest range of Z parameter values, we set the cluster size N=100.

**Figure 5** shows that increasing the parameter Z lowers the likelihood that any individual is at risk of being privacy exposed i.e. increases the effective privacy. For example, a y-axis value of 0.1 means that there is a 90% chance that no individual is at risk of being privacy exposed, and thus there is at least a 90% chance that there is no possible deductive reasoning to infer the membership of any individual in the input dataset. We also overlay the sliced-Wasserstein distance between the real and synthetic dataset, visualising the trade-off between accuracy and privacy, which is further illustrated by a PCA plot for Z=0 versus Z=6 (**Supplemental Figure 14** and **15**). From this figure, Genomator provides a 20 times improvement in privacy (1 to 0.05) when sacrificing 33% accuracy (2% vs 3%).

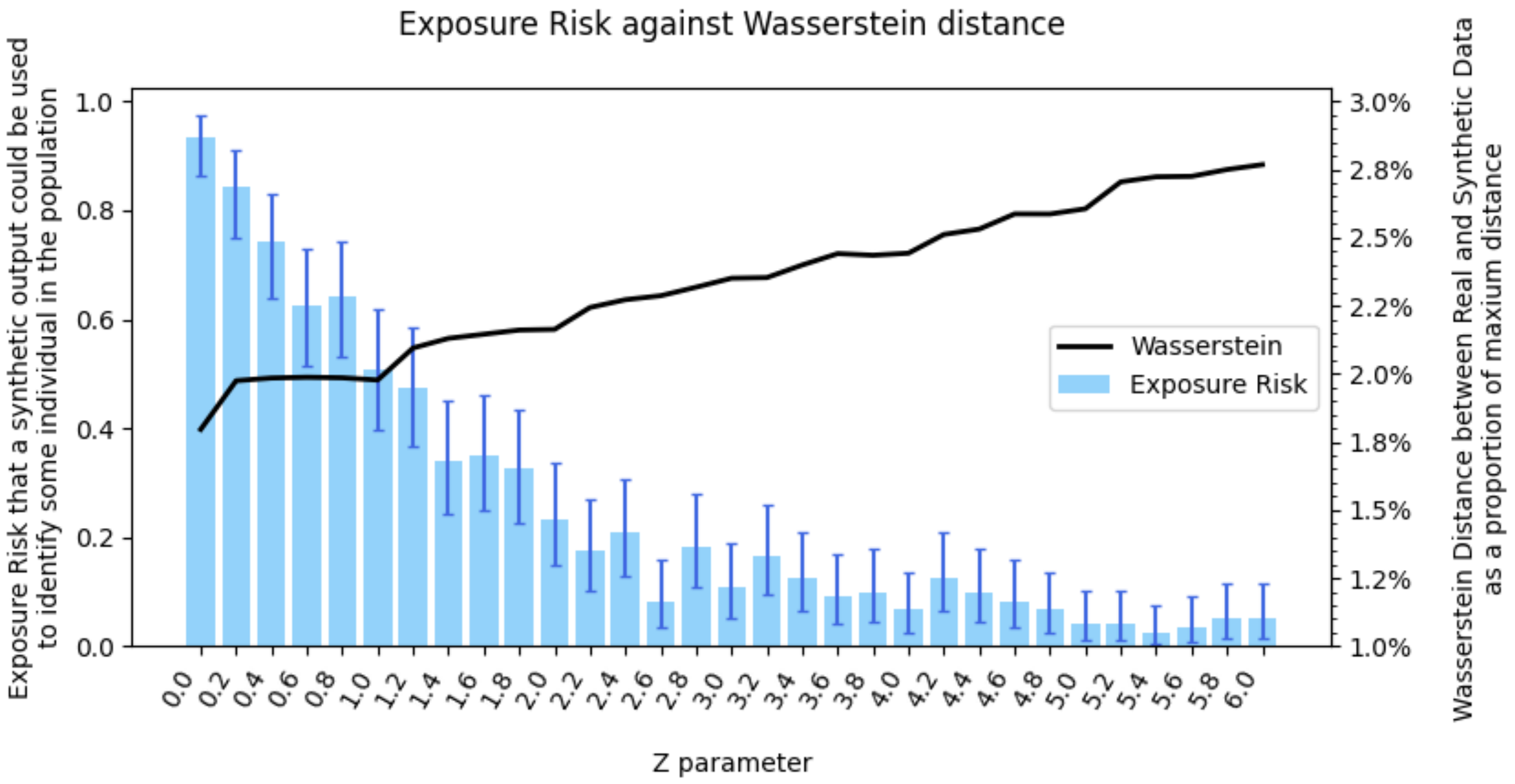

Figure 5: Plot showing the trade-off between privacy and accuracy for Genomator. Shown in blue the experimental likelihood that a synthetic output from Genomator could be used by Reverse Genomator to logically identify an individual in the input dataset (Bernoulli distributed, 90% confidence intervals are shown), across Z parameter values for Genomator. Shown in black the Wasserstein distance between synthetic data generated by Genomator for those parameters and the real dataset.

To drill down into the privacy dynamic further, we quantify how much information is likely revealed about a target individual in the context of the synthetic output from the input dataset in which they are present. Particularly, we quantify how often "private" SNP combinations leak into the synthetic data output. These are sets of SNPs that are only seen in one individual in the dataset and hence replicating them might expose uniquely identifiable features of that individual and facilitate identification. We compare this to the "hallucination" rate at which fictious SNP combinations (that are not seen in any individual) are created to quantify the balance between accuracy and privacy.

We generated 1000 synthetic versions of the *AGBL4* gene using each of the methods and computed the likelihood that private and fictitious combinations of SNPs appeared in the output dataset as outlined in **Methods 2.4**. The ideal method produces fewer privacy-revealing combinations, while avoiding creating unseen combinations that are potentially not viable in humans.

**Figure 6** shows that Genomator is significantly more hallucination-resistant than GAN, RBM, CRBM, and Markov (t-test *p*=0.0005). Genomator is the only method able to operate in the optimal quadrant with 95-98% better privacy and 132-388 times better hallucination-resistance compared to GANs**,** RBM, CRBM, and Markov, respectively compared on method averages (see **Supplemental Table 7**).

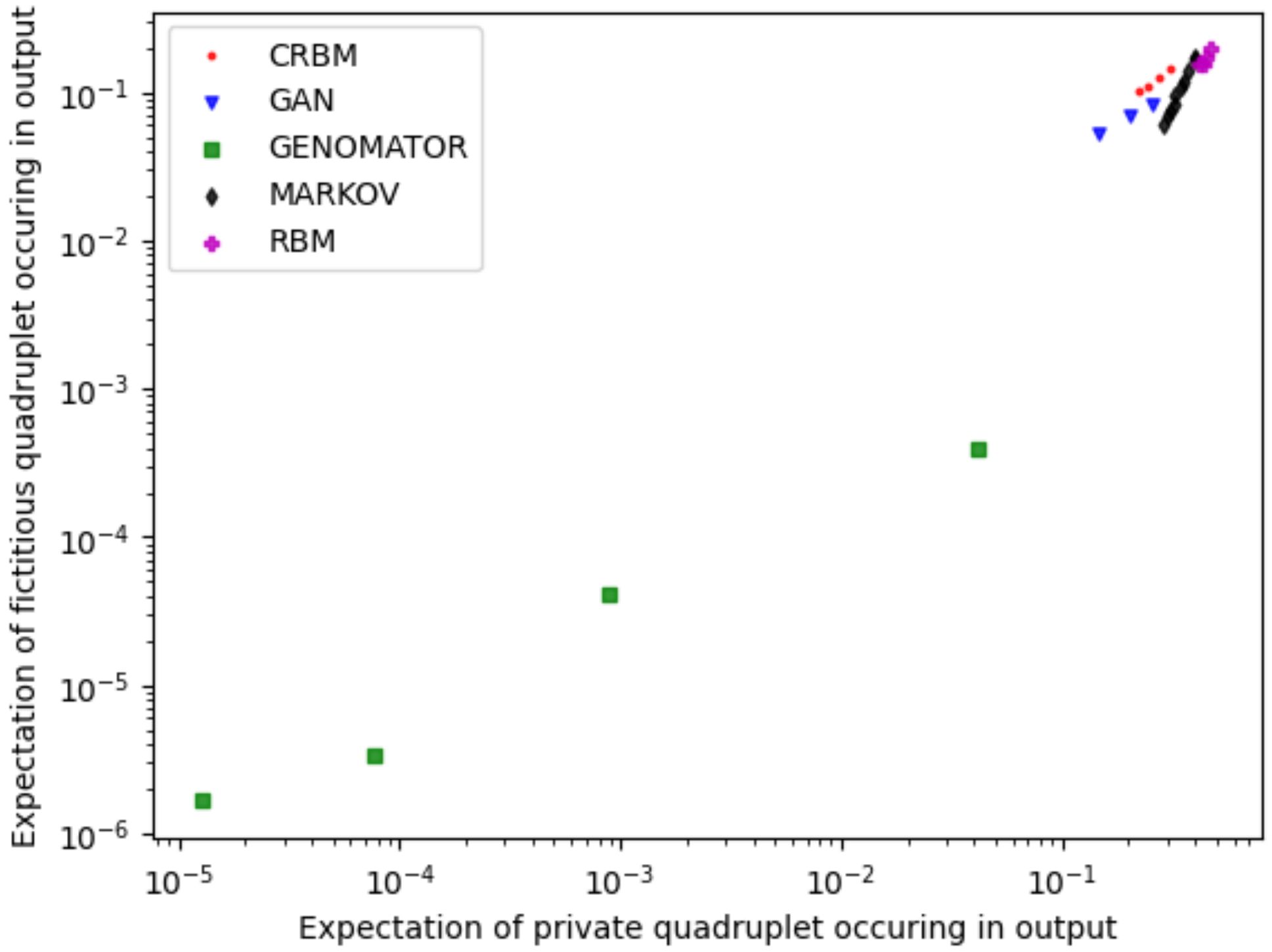

Figure 6: For each method, the likelihood of quadruplet combination of SNPs in the output being 'fictitious' vs private. Where a fictitious quadruplet is one that is not featured in the input dataset and a private quadruplet is one that is featured in exactly one individual in the input dataset. Each of the mechanisms were run with a series of parameters against Genomator – Where Genomator is run from N=10->25 together with Z=0->3 in unit increments. For a septuplet evaluation see Supplemental Figure 13.

### 3.5 Utility: Reproduction of well-known pharmacogenetic SNPs in large synthetic datasets

Genomator was used to create a synthetic dataset of 1000 individuals across the 1.11M SNPs of chromosomes 10 and 16. Creating whole chromosomes was feasible with the available resources for Genomator but would have been impossible for the other methods as shown in **Section 3.2**. We hence omitted them from this analysis. Chromosomes 10 and 16 were chosen as they include three important pharmacogenetic genes and SNPs, namely *CYP2C19* (*rs*4244285 and *rs*4986893) and *VKORC1* (*rs*9923231). Genetic variants in the *CYP2C19* gene have been associated with differential metabolism of the antiplatelet drug clopidogrel, where individuals with two loss of function (LOF) alleles have significantly reduced response to the drug[36]. Importantly, the most studied and common LOF *CYP2C19* variants, *CYP2C19*2* (*rs*4244285) and *CYPC129*3* (*rs*4986893) have significant varying allele frequencies between ethnicities where both LOF alleles are more frequent in individuals of East Asian compared to those of European and African descents[37]. Similarly, the SNP *rs*9923231 in the *VKORC1* gene is significantly associated with increased warfarin sensitivity and explains the differences of warfarin dosage requirements between ethnicities[38].

Genomator generated samples clustered into four super-population groups based on PCA projection with the 1KGP reference samples: African (AFR), East Asian (EAS), European (EUR), and Others (**Supplemental Figure 10, Supplemental Table 6**). The calculated minor allele frequencies of the three important pharmacogenetic SNPs from Genomator's generated data were comparable to those of three reference datasets: the 1KGP, gnomAD v4, and the Human Genome Diversity Project (HGDP) across the ethnicities (**Figure 7**).

This result shows that Genomator can emulate clinically relevant population-specific allele frequencies at the SNP-level across chromosomes, as well as preserve local genomic structures (LD) and population-specific SNP-frequencies across the whole genome (PCA) as shown in **Section 3.1**.

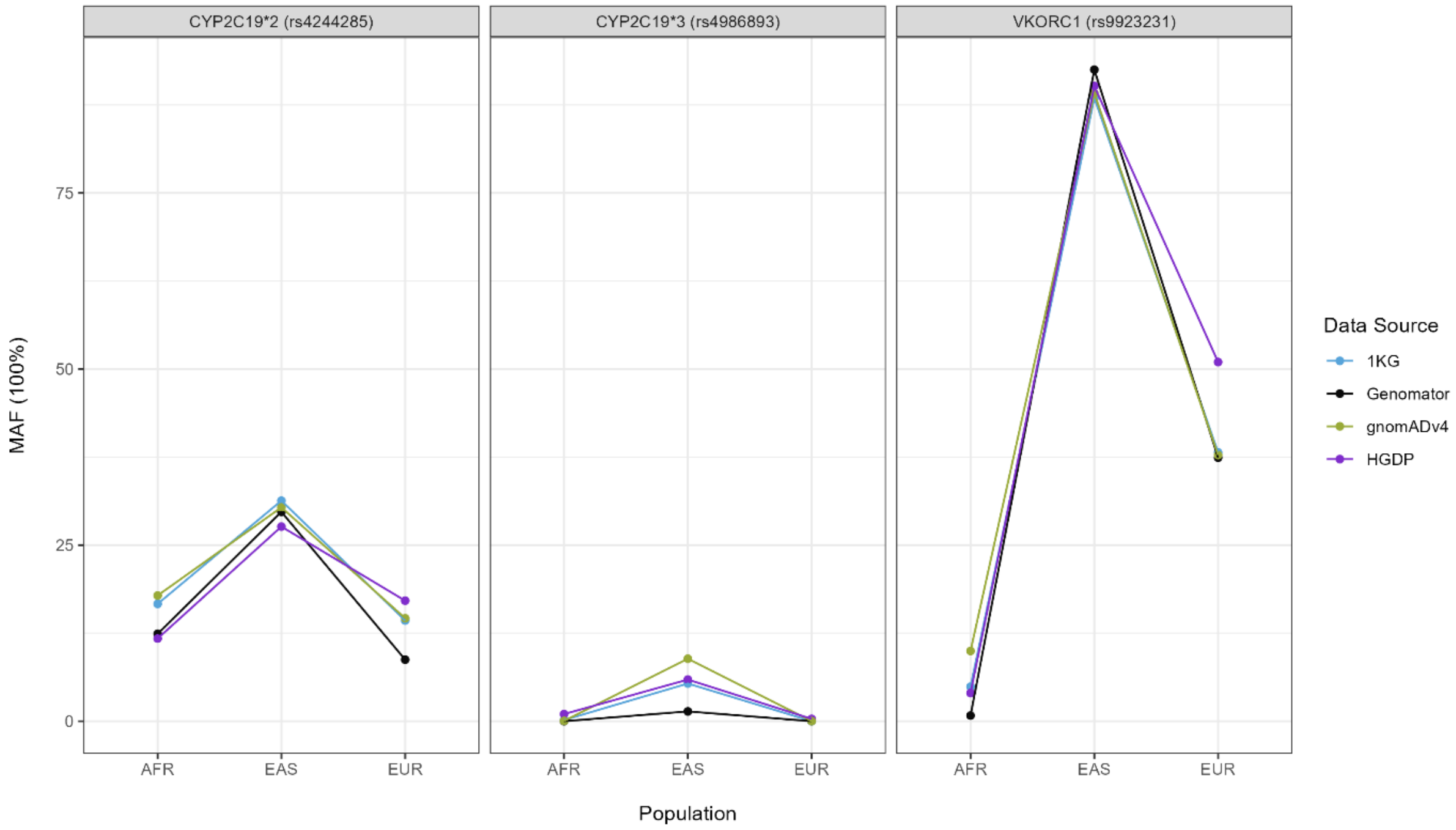

Figure 7: Plot comparing minor allele frequencies of three important pharmacogenetic SNPs (*rs*4244285, *rs*4986893, *rs*9923231) calculated from four data sources: (1) 1000 Genomes Projects (1KG) (*n* = 1,668), (2) Genomator generated samples (*n* = 637), (3) gnomAD v4 (*n* = 682,050), (4) Human Genome Diversity Project (HGDP) (*n* = 526), across three ethnicities (African/African America (AFR), East Asian (EAS), European (EUR)).

## 4. Discussion

While all methods were able to produce genomes that broadly reflect the input data, Genomator more effectively captures secondary statistics, such as gene-gene correlations (i.e. LD) because it reasons directly over all pairwise associations in its generation process. By contrast, GAN and RBM are limited in the amount of information that they can reproduce by the size of their network and hidden layers, and Markov chains are bounded by the chosen haplotype window size in capturing higher-order genome-wide structures. Genomator has an 84-93% reduction in LD error compared to other tools (0.000206 vs 0.00132-0.00304, **Supplemental Table 2a**). CRBM was observed to perform worse than RBM owing to having fewer connections between visible and hidden layers and featuring conditioned sequential generation in a scan across genome positions.

Genomator can reason over all pairwise associations across the millions of SNPs in the human genome, by the efficiency of two variable constraints in the SAT solving process see **Supplemental Section 1**. Reasoning in such a way means that the constrainedness of the SAT problem increases with dimensionality. Unlike other tested methods, where capacity for accuracy degrades as data dimensionality grows, Genomator becomes more accurate as the scale of the input data increases. This is because more SNPs means there are more pairwise signatures to compare and more constraints to put into the SAT solver, which further constrains the logical deduction to generate genomes that are similar to the input, resulting in higher accuracy. In this context, Genomator can create whole genomes using a single-thread CPU and minimal RAM, 1000-1600 times faster at scale (4.95 vs 5283.01-8383.07 seconds tested at 1,638,400 SNPs as handled by most methods, **Supplemental Table 5**), generating a full synthetic human genome in 18 seconds.

While synthetic data generation methods can be evaluated for privacy by proxy, to date only Genomator offers an approach to deduce privacy (**Supplemental Figure 3**) and hence precision-tune the accuracy-privacy trade-off to the individual application needs. This shows a 95%-98% privacy improvement (average of 0.01 vs 0.2-0.4 private quadruples leaked, **Supplemental Table 7**) over other methods and can be tuned over 3000-fold (privacy exposure from 0.041 to 1.3E-05, **Supplemental Table 7**)). Furthermore, while adding privacy conservation can in theory be built into the generation process for GANs[39], having explicit "diversity" clauses in conjunction with the privacy parameter Z, ensures Genomator does not leak rare SNP combinations. This is especially difficult for generative models where growing datasets exposes them to more genetic diversity during training, which in turn increases the chance for rare SNPs to be replicated in the output. With no parametric control during the generation phase for those methods, privacy violating SNPs will have to be removed retrospectively, which creates obvious "holes" in the data from which information can also be deduced.

We demonstrate the potential in making cohorts accessible to research and the clinic, by creating a representation of a global data resource that contains population-specific information capable of guiding, for example, pharmacogenetic algorithms for dosage levels of warfarin. It is important to note that ultra-rare variants can have clinically actionable effects and may be important to include in synthetic datasets. However, Genomator attenuates ultra-rare SNPs present in the input dataset, as can be observed when replicating *rs*4986893 in the EAS population (**Figure 6**). While problematic for replicating ultra-rare variants, this tendency is overall beneficial as it protects against privacy violations for individuals with such *private* mutations. The sharing of ultra-rare genomic events under privacy considerations hence remains an unsolved issue.

We note that the use of synthetic genomes may be subject to legal and ethical constraints depending on where and how they are used or shared. The requirements and legal responsibilities around synthetic genomes, such as identifying appropriate risk minimisation [40], and how technology and techniques may interact with applicable regulations and laws[41] may need to be considered.

Furthermore, the American College of Medical Genetics and Genomics (ACMG) [42] recommends that significant findings are shared with the participant. To do this, the synthetic sample would need to be traced back to the individual it originated from. This concept is known as "Pseudonymous data", which allows to re-identify a person, but only when combined with other information. Genomator is the only tool that can meet this recommendation by running Reverse Genomator with the used clauses over the original dataset to identify the individual or small set of individuals the sample has originated from. Please note, this does not represent a vulnerability to leaking data as only the data owner who has full access to the cohort can perform a re-identification.

While Genomator has a lower resource footprint than other approaches for processing inputs, it can further benefit from multiple cores and/or GPU implementation. Especially when creating thousands of genomes where Genomator has to process the logical constraints for each genome anew, whereas trained deep learning-based generative models can simply generate more outputs. Additionally, Genomator's SAT constraints can be extended, such as by adding constraints to avoid leaking disease status, or modified completely to cater to data applications outside the genomic or medical domain.

**5. Conclusion**

In this study we presented Genomator and demonstrated that it is possible to generate synthetic data using a SAT solver to reproduce genomic information. Our technique is accurate, scalable, and computationally efficient as well as configurable to retain the genomic privacy of the individuals in the source dataset. Additionally, we have developed Reverse Genomator to deductively and logically inspect the output from Genomator and calculate the absolute privacy afforded by Genomator.

The use of private synthetic data in lieu of real data may allow institutions and biobanks to share genomic information more liberally and advance health applications and knowledge.

**Acknowledgements:** We would like to acknowledge Associate Professor Charles Gretton for inspiring our SAT solving enthusiasm.

**Author contributions.** MB conceived the approach and conducted the experiments with contributions from RR, AK, and MOB. LMFS conducted the pharmacogenomics experiments. BH made the work reproducible. YJ and DCB supervised the approach and shaped the paper. All authors contributed to the final document.

**Competing interests:** Provisional Patent filed (29967057)

**Funding:** This work has been supported by CSIRO.

**Conflict of Interest:** none declared.

**Availability**: Scripts to reproduce experimental data is available at https://github.com/aehrc/genomator and a Code Ocean capsule for the paper is being setup and will be available via the manuscript system.

**Supplemental for: Privacy-hardened and hallucination-resistant synthetic data generation with logic-solvers**

Mark A. Burgess[1], Brendan Hosking[2], Roc Reguant[2], Anubhav Kaphle[3], Mitchell O'Brien[2], Letitia M.F. Sng[2], Yatish Jain[2,5], Denis C. Bauer[4,5]*

[1] Australian e-Health Research Centre, Commonwealth Scientific and Industrial Research Organisation, Canberra
[2] Australian e-Health Research Centre, Commonwealth Scientific and Industrial Research Organisation, Sydney
[3] Australian e-Health Research Centre, Commonwealth Scientific and Industrial Research Organisation, Melbourne
[4] Australian e-Health Research Centre, Commonwealth Scientific and Industrial Research Organisation, Adelaide
[5] Macquarie University, Applied BioSciences, Faculty of Science and Engineering, Macquarie Park, Australia

# Supplemental Section 1: Detailed description of Genomator Algorithm

## Section 1a: SAT solving background

The Boolean Satisfiability Problem (SAT problem) is the problem of determining if there is a set of values which can be assigned to a range of binary variables such as to make a particular Boolean formula evaluate to being True. If there is such a valuation, then the problem is said to be 'satisfiable', or otherwise 'unsatisfiable'.
The set of Boolean variables is usually finite, and denoted with indices such as: x1, ..., xM (where M is the number of variables in the formula).
The formula is often constructed in Conjunctive Normal Form (CNF) as a logical conjunction of disjunctions of literals. A literal is an expression of a binary variable or its negation, denoted 'x' and '¬x' respectively (symbol '¬' is logical negation). A disjunction of literals is called a 'clause', such that 'x1 OR ¬x3 OR x5' is an example of a clause - which is an expression that is True if any one of its literals evaluates to True, and False otherwise (a clause with three literals is considered to have a size of three. etc). A conjunction of clauses makes a formula in CNF: e.g. '(x1 AND ¬x3 OR x5) AND (x3 OR ¬x1)' which is True if all of the clauses evaluate to being True, otherwise False.
Formulas are often input into SAT solvers in CNF form, and a SAT solver will attempt to determine values to all the variables such that the formula evaluates to True.
We note that many different binary formulas can be constructed to represent a range of problems in computer science and these formulas can be converted into CNF form by application of De Morgan's laws.
There are a range of SAT solvers which are available, and each behave slightly differently in operation. Particularly we are concerned with using a class of SAT solvers called 'backtracking' SAT solvers, which attempt to solve a SAT problem systematically, by iteratively assigning values to binary variables and systematically backtracking in order to search for a solution.
The Genomator software uses the PySAT library[1] to interface with a number of available SAT solvers. The background and features of SAT problems and SAT solving approaches is well documented[2].

## Section 1b: Concepts

The central rationale of Genomator's algorithm is that it considers all pairs of simple features that none of the inputs data possess, and generates synthetic data simply by ensuring that these pairs of features do not occur in the synthetic data. This principle motivates the elements of the design and process of the algorithm.

A SAT solver is used for this purpose, where a set of binary variables are considered which determines the output synthetic data, and a scan is tacitly done over all simple features of the input data, and where there is a pair of such features that does not exist across the input data, constraints are added to the SAT solver to prohibit such a pair of features from being in the output synthetic data.

The entire flowchart of the Algorithm (with its pseudocode) is given as **Supplemental Figure 1** (and **Supplemental Figure 2**) respectively. And in the following subsections we will break down the parts of this algorithm and describe them (with their motivation) in turn.

## Section 1c: Input Dataset

We consider aligned genomic information together with the yet-undetermined synthetic data (denoted by blue '?' characters), such as shown in **Supplemental Figure 1a**. For simplicity of illustration we consider that there is a single nucleotide base (A,T,G,C) at a location.

| | Person 0: | Person 1: | Person 2: | Person 3: | Person 4: | Synthetic: |
|---|---|---|---|---|---|---|
| Location 1: | A | T | C | A | A | ? |
| Location 2: | T | G | A | T | T | ? |
| Location 3: | G | C | T | G | C | ? |
| Location 4: | C | A | G | C | G | ? |
| Location 5: | T | G | G | G | T | ? |

Supplemental Figure 1a: Aligned genome information, illustratively showing a single nucleotide base at each genomic location together with synthetic data – denoted by '?' characters which are yet to be determined by the algorithm. For simplicity of illustration purposes, we show single nucleotides across genome locations, but the algorithm can also process diploid genomes.

In this context, Genomator considers the presence and absence of nucleotides across locations as simple features – such as answered by queries such as "do you have an A at position 4?", as well as "do you NOT have an A at position 4?", "do you have an T at position 2?", as well as "do you NOT have an T at position 2?", etc.
And where some pair of these features does not exist in any input datum, the synthetic data output is prohibited from possessing that pair of features.
For instance, in the above dataset, there is no input datum which has a G at location 3 and does not have an A at position 1; and therefore, the synthetic data will be similarly so.

This form of processing has some natural consequences: for instance, the more data (and the more diverse the data) that is input into Genomator, the more likely there are to be an instances of pairs of features, and the less constrained the synthetic data output will be by this process. Or conversely, the fewer the input data (or the less diverse) the more likely that there is to be no instances of pairs of features across the data, and the more constrained the synthetic data output will be. The number of input data – ‘N’ - is an important parameter in the method.
Another observation about this processing is that the greater the scale of the data (that is the more locations there are in the data) the more pairs of features there are to consider and the more numerous and thus restrictive the combinations of prohibited feature pairs become, conversely the smaller the scale of the data (that is less locations) the less pairs of features there are to consider and there is a smaller mass of restricting implication between them.

A modification we also consider is rather than only prohibiting pairs of features in the synthetic data that belong to none of the input data, we can also consider prohibiting pairs of features in the synthetic data that belong to exactly one input datum, also two input datum, or more – a parameter we denote with ‘Z’.
In this way, if a pair of features appear in the synthetic data output it must therefore have been present in at least Z of the inputs.
Furthermore we randomise this ‘Z’ value across pairs of features considered – and this randomised parameter ‘Z’ is an important privacy parameter in the method as it directly (and/or randomly) excludes rare pairs of features from appearing in the synthetic output that could be privacy revealing of particular inputs.

The process of selecting a number of inputs ‘N’ is considered in the pseudocode (**Supplemental Figure 2**) on line 3, and the process of drawing a random ‘Z’ per pair of positions is indicated on 18.

## Section 1d: Processing elements

The First stage of generating synthetic data by our method consists of considering all simple features in turn, and generating a set of binary arrays (or ‘signatures’) holding whether each datum satisfies the respective query. As shown in **Supplemental Figure 1b**.

```
             EACH QUERY (AND ITS NEGATIVE) AT EACH LOCATION
                     GIVES A TRUE/FALSE SIGNATURE

         -      Are you a A : T F F T T ?
Location -      Are you a G : F F F F F ?
         -      Are you a T : F T F F F ?
    1    -      Are you a C : F F T F F ?
         -  Are you NOT a A : F T T F F ?
         -  Are you NOT a G : T T T T T ?
         -  Are you NOT a T : T F T T T ?
_________-  Are you NOT a C : T T F T T ?
         -      Are you a A : F F T F F ?
Location -      Are you a G : F T F F F ?
         -      Are you a T : T F F T T ?
    2    -      Are you a C : F F F F F ?
         -  Are you NOT a A : T T F T T ?
         -  Are you NOT a G : T F T T T ?
         -  Are you NOT a T : F T T F F ?
_________-  Are you NOT a C : T T T T T ?
         -      ... ... . X : : : : : : :
         -      ... ... . X : : : : : : :
         -      ... ... . X : : : : : : :
         -      ... ... . X : : : : : : :
```

Supplemental Figure 1b: A section of the ‘signatures’ associated with the aligned genomic information (shown in Supplemental Figure 1a) which are binary arrays –showing True/False- for each query (positive and negative) about possession of a specific nucleotide at each location. Also showing the undetermined synthetic response to these queries denoted by blue ‘?’ characters.

In this figure we notice that every query (eg. “at location 1, are you a A?”) has a signature, such as (T, F, F, T, T) indicating which of the input data satisfy the condition (or not). Additionally we depict that the synthetic data response to the query is yet undetermined – shown adjacent as blue ‘?’ characters – which might True or False.
We note that the synthetic data responses (shown as blue ‘?’ characters) can be used to determine what the synthetic data nucleotides are.
For instance, If the synthetic data response to query “Are you an A at location 1?” is True, then this is sufficient to determine one nucleotide of the synthetic data.
This inference is unambiguous if the answer to all similar queries “Are you a G at location 1?” (or a C, or T) are False, and the inverse queries “Are you NOT an A at location 1?” (and for G/T/C) are oppositely responded to their non-inverted counterparts.
Genomator uses a SAT solver to derive the synthetic data responses to all the queries, which then determines the synthetic data entirely.

The SAT solver functions based on an input set of constraints, and so we need to consider what constraints should be added to generate unambiguous synthetic data.
To determine these constraints, we note that as the synthetic data must have exactly one nucleotide (A/T/G/C) at each location, that therefore there must be exactly one synthetic response that is ‘True’ to the queries “Are you an A?”/T/G/C at each location, notably this means that there that there must be at-least-one ‘True’, and that there cannot be any two that are ‘True’.
We will subsequently see that the first of these constraints will be added explicitly (**Supplemental Section 1g**), the second will be handled indirectly (**Supplemental Section 1f**).
Additionally, the constraint that the synthetic response to inverted queries must be opposite will be handled automatically by determining them to be opposite values of specific binary variables in the SAT solving process (**Supplemental Section 1e**).
These constraints are sufficient to ensure that there is exactly one unambiguous way that synthetic genome data can be read from the synthetic data responses to the queries (which is explained in **Supplemental Section 1h**).

The process of converting aligned genome information into signatures via a series of queries is considered in the pseudocode (**Supplemental Figure 2**) on lines 6,7,8 – where for each nucleotide A/T/G/C variable, Q and W Boolean arrays are generated where the selected genomes match the respective nucleotide.

Both positive and negative queries are considered, as this strengthens the logic employed by Genomator, as per the rationale of Genomator (see **Supplemental Section 1b**) all features of the data (both positive and negative) are considered, and where a pair of such features is not present in the input data, they are not reproduced in the synthetic data.

## Section 1e: Deduplicating Signatures

In the previous subsection, we considered that each query had an associated signature, and that the synthetic response to these queries determined the synthetic genome data.

In this context we know that the synthetic data response to queries with identical signatures must be identical. And this is a direct consequence of the central rationale of Genomator's algorithm – see **Supplemental Section 1b**.

To see this, consider the queries (such as shown in **Supplemental Figure 1b**), such as "Are you an A at location 1?" and "Are you a T at location 2?" have identical signatures (T F F T T), if the synthetic data response to these queries were not identical – such as having an A at location 1 but not having a T at location 2 (or otherwise) then the synthetic data would have a pair of features that none of the input data possess (see **Supplemental Figure 1a**) and such a possibility is therefore to be excluded.

Because of this reasoning, we can consider that the synthetic data response to queries of identical signatures should be identical, and therefore we can turn to focus on responses over signatures that have the potential to be different.

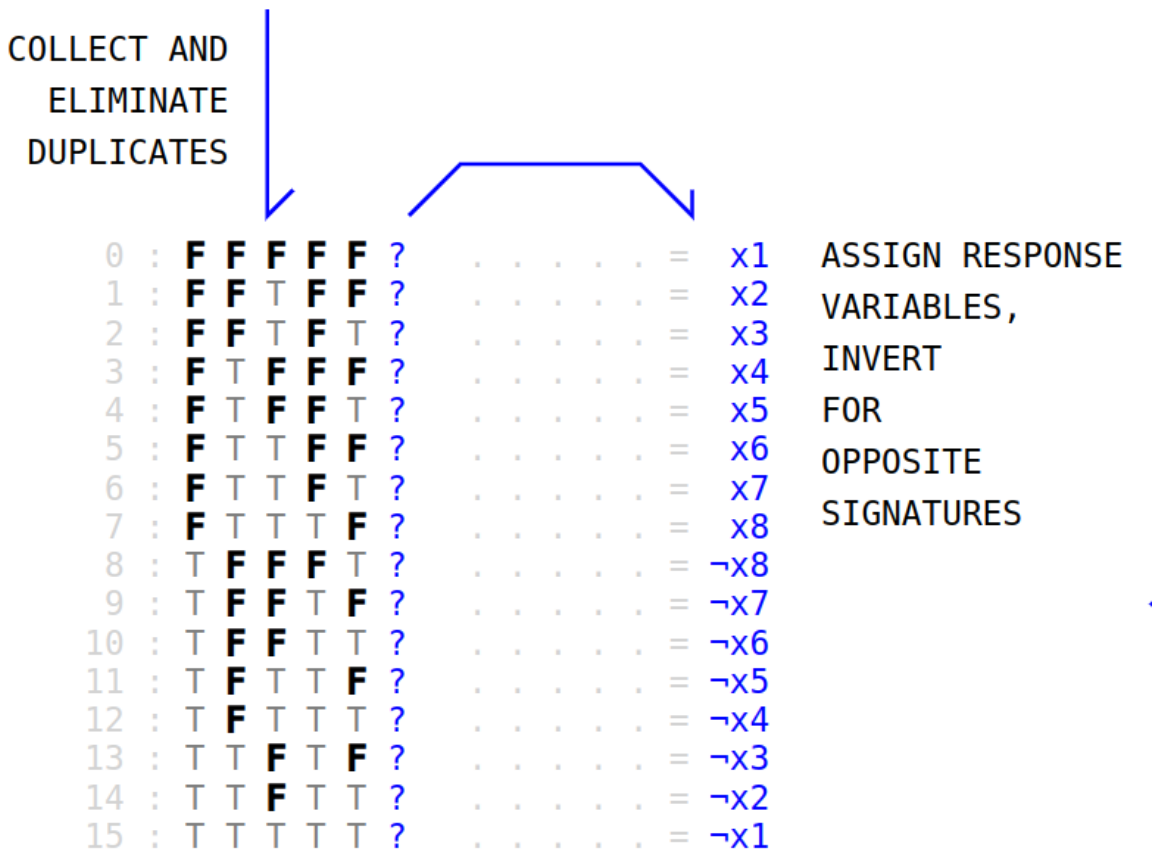

Supplemental Figure 1c: A deduplication of all the signatures created by Genomator over the input genome dataset (see Supplemental Figure 1b) where the synthetic response to queries associated with these signatures is unknown (denoted in blue '?' characters), these are then assigned to be the value of binary variables, such that each unique signature is associated with a unique literal, and inverted literals are associated with inverse signatures. The unique signatures are indexed 0-15.

Consider **Supplemental Figure 1c**, where we have taken the signatures depicted in **Supplemental Figure 1b** and have deduplicated and sorted them by their indices (first column, second column, and so on) to form an ordered list of unique signatures.
Each of these unique signatures have the potential to be associated with different synthetic responses, and so we construct the synthetic response to each to be given by the value of binary variables in the SAT problem where each binary variable indicates the synthetic data response across all queries with the same signature.
For instance, the inverted binary variable value ¬x6 is associated with signature (T F F T T) identifying the synthetic response to all queries of that signature (such as "Are you an A at location 1?" and "Are you a T at location 2?" etc.).

As it is identified in **Supplemental Section 1d** that the synthetic response to inverted queries must be opposite (for the resolved synthetic data to be unambiguous), this means that the synthetic response associated with inverted signatures must be opposite.
For instance: the synthetic response to query "are you an A at location 1?" must be opposite to the query "are you NOT an A at location 1?", therefore the synthetic response to all queries with signature (T F F T T) must be opposite to that of signature (F T T F F).

One of the simplest ways of encoding this feature is employed in our construction, which is to have signatures and their inverse (such as (T F F T T) and (F T T F F)) associated with binary variables and their inverses (in this case ¬x6 and x6 respectively). This directly ensures that queries that yield opposite signatures (such as “are you an A at position 1?” and “are you NOT an A at position 1?”) are responded oppositely – which would be senseless otherwise.

The deduplication and sorting of the signatures – such as depicted in **Supplemental Figure 1c** – is done to order the signatures in such a way to readily distinguish all inverted pairs.

The deduplication of signatures and the assigning of binary variables to these signatures is considered in the pseudocode of the algorithm (**Supplemental Figure 2**) in lines 9 and lines 11-14.
Where on line 9, the sets of signatures Q and W are unioned with the set ‘signatures’ (thereby duplicate signatures are removed).
On lines 11 and 12 – the maximum number of binary variables ‘n’ is calculated (as half the number of unique signatures), the set of signatures is sorted (first column, then second column, and so on – mirroring that shown in **Supplemental Figure 1c**)
On lines 13 and 14 – the binary variables are ordered (x1 to xn and ¬xn to ¬x1) such as to be matched in order with the sorted binary signatures (see **Supplemental Figure 1c**), and the function f(s) is declared to return the binary variable associated with an input signature ‘s’.

Now that we have associated binary variables (and inverse binary variables) with signatures, we construct constraints for the SAT solver using these variables. Particularly, such as to ensure that all pairs of features that the dataset do not possess do not occur in the synthetic data (see **Supplemental Section 1b**), this requires some enumeration of such feature pairs – which we consider next.

## Section 1f: Pairing Unique Signatures

In the previous subsection we considered that queries with identical signatures should be responded to by the synthetic data identically, however this is not quite sufficient to guarantee that all pairs of features which do not occur in the input data are not replicated in the synthetic data. To ensure this we need to enumerate over pairs of signatures.

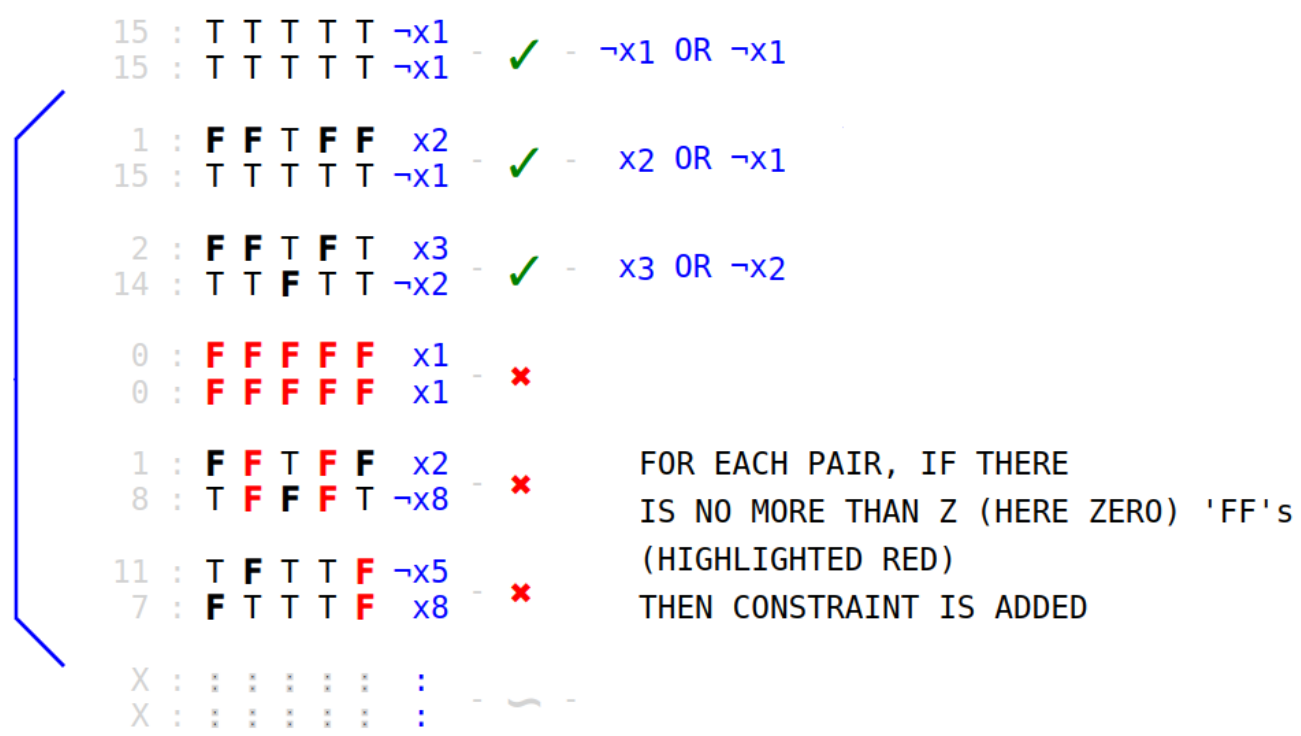

Supplemental Figure 1d: A section of the pairs of unique 'signatures' created by Genomator, and an illustration of the method of transforming these pairs into relevant logical constraints for input into the SAT solver, based on the prominence of 'False-False' pairs across signature positions.

In **Supplemental Figure 1c** we enumerated unique signatures and indexed them from 0 to 15, and in **Supplemental Figure 1d** we show some pairs of these signatures (such as indexed 15 and 15, 1 and 15, 2 and 14, 0 and 0, etc.) we note that doubles (such as index 15 and 15) are included in the analysis.

To ensure that pairs of features that are not part of the input data are not reproduced in the synthetic data, we consider all pairs of signatures and look for the absence of a specific relationship in each – in our case we focus on a 'False-False' relationships specifically – though we will see that this is arbitrary and occurs without loss of generality.

Some pairings are depicted in **Supplemental Figure 1d**, such as between pairs indexed 1 and 15 – F F T F F and T T T T T. In this context there is an absence of a False-False relationship between the positions of these signatures (shown in the **Supplemental Figure 1d** in red) - as the relationships between these signatures across positions are: False-True, False-True, True-True, False-True and then False-True.
Because there is an absence of any False-False relationship, the synthetic data responses (given by binary variables x2 and ¬x1) cannot be False-False, and this is ensured by adding the constraint 'x2 OR ¬x1' to the SAT solver (which ensures that x2 and/or ¬x1 must be true). This logic is depicted in **Supplemental Figure 1d** between a range of pairs, and where there does not exist a False-False relationship the corresponding constraint is added to the SAT solver.

This logic ensures that the SAT solver assigns values to all binary variables such that pairs of features that the input data do not possess, are not reproduced in the synthetic data.
And this tacitly captures some degenerate cases, for instance, no input datum has both an 'A' nucleotide and also a 'T' nucleotide in the same location, and consequently these constraints ensure that the values of the binary variables will not indicate that the synthetic data has both an 'A' and a 'T' nucleotide at any same location.
This logic holds across all pairs of nucleotides and across all locations, and is one part of ensuring that the values to binary variables encode a sensible and uniquely readable synthetic genome (see **Supplemental Section 1d**) – we note that this complements the addition of at-least-one constraints (explicitly addressed in the next **Supplemental Section 1g**) to ensure that there is exactly-one A/T/G/C at every location.

Breaking this down: the query “are you an A at location 1” is associated with signature T F F T T, and the synthetic response to this query is given by ¬x6, and the query “are you a G at location 1” is associated with signature F F F F F, and the synthetic response to this query is given by x1 (see previous **Supplemental Figures 1b** and **1c**). In this context the constraint that the synthetic data cannot be both an A and a G at location 1 is naturally given by constraint: ‘NOT (¬x6 AND x1)’ or equivalently ‘x6 OR ¬x1’. This constraint is formed by the algorithm at the junction where it considers the pair of these inverted signatures F T T F F and T T T T T which are associated with x6 and ¬x1, and because there is no False-False relationship between the positions of these signatures the constraint ‘x6 OR ¬x1’ is appropriately added.

As another instance (as foreshadowed in **Supplemental Section 1b**), there is no input datum (shown in **Supplemental Figure 1a**) which has a G at location 3 and does not have an A at position 1, and therefore the synthetic data will be similarly so.
The query “Are you a G at location 3” is associated with signature T F F T F and synthetic response is given by ¬x7; the query “Are you not an A at position 1” is associated with signature F T T F F and synthetic response is given by x6. As no input datum possesses a G at location 3 and does not possess an A at location 1, the appropriate constraint naturally is: ‘NOT (¬x7 AND x6)’ or equivalently ‘x7 OR ¬x6’. Which is added when the algorithm realises there is no False-False relationship between positions of the inverse signatures F T T F T and T F F T T associated with x7 and ¬x6 respectively.

To see that considering the False-False relationship is sufficient and without a loss of generality, we note that we could also consider that in the pairing of signatures indexed 1 and 15 (shown in **Supplemental Figure 1d**), there is also the absence of a True-False relationship between the positions of these signatures. Because there is an absence of any True-False relationship, the synthetic data responses (given by binary variables x2 and ¬x1) cannot be True-False, and this would be ensured by adding the constraint ‘¬x2 OR ¬x1’ to the SAT solver (which ensures that ¬x2 and/or ¬x1 must be true). However this constraint would arise just as much from focusing on False-False relationship and the pair between index 14 and 15 – T T F T T and T T T T T – where the absence of a False-False relationship would yield the same constraint ‘¬x2 OR ¬x1’ to the SAT solver.
In this way, because all signatures have an inverse, and their inverse is associated with an inverted binary variable, considering all pairs of signatures and focusing only on False-False relationships is sufficient to generate all these relevant constraints.

We noted earlier (in **Supplemental Section 1c**) that we also consider a modification to this central logic, whereby a pair of features is excluded from being a part of the synthetic data if it occurs zero, or one, or two times – up to a parameter ‘Z’ - in the input data etc.
This parameter ‘Z’ becomes relevant at this junction of considering pairs of signatures and evaluating a threshold level of many False-False relationships exist between the positions of paired signatures – as illustrated in **Supplemental Figure 1d**.
If ‘Z’ is zero, then if there are no instances of a feature pair in the input dataset, and there are no False-False relationships in the relevant signature pair, and consequently a constraint is added which prohibits that feature pair in the synthetic dataset.
If ‘Z’ is one, then if there are one-or-less instances of a feature pair in the input dataset, then there is one-or-less False-False relationships in the relevant signature pair, and consequently a constraint is added that prohibits that feature pair in the synthetic dataset. ...and so on.

The process of comparing unique signatures and forming constraints from them is considered in the pseudocode of the algorithm (**Supplemental figure 2**) on lines 17-20. On line 17, all pairs of signatures are compared in a 'for' loop.
On line 18, a random number 'z' is drawn from a probability distribution 'Z'
On line 19, the number of 'False-False' pairs are calculated by OR'ing the signatures together per location, and if less than 'z' 'False-False' pairs exist then, on line 20, a size-2 clause is added to the SAT solver constituent of the two binary variables associated with those signatures.

This signature-pair processing prohibits simple feature pairs that do not exist in the input data (but for 'Z' instances) from appearing in the synthetic data.
This processing includes the degenerate case of prohibiting the binary variables of the SAT problem from indicating that the synthetic data has multiple nucleotides at a particular location. However this is not quite sufficient to ensure that there is exactly one nucleotide at a particular location – which is delt with next.

## Section 1g: Adding at least-one constraints

In the previous section we considered constraints that prohibit pairs of features that do not exist in the input data from being replicated in the synthetic data, notably this includes constraints that ensure that values are assigned to binary variables such as to indicate that no two nucleotides are simultaneously present at any particular location. However this is not quite sufficient to ensure that values are assigned to binary variables to indicate that there is some nucleotide at every location (ie. Not none).

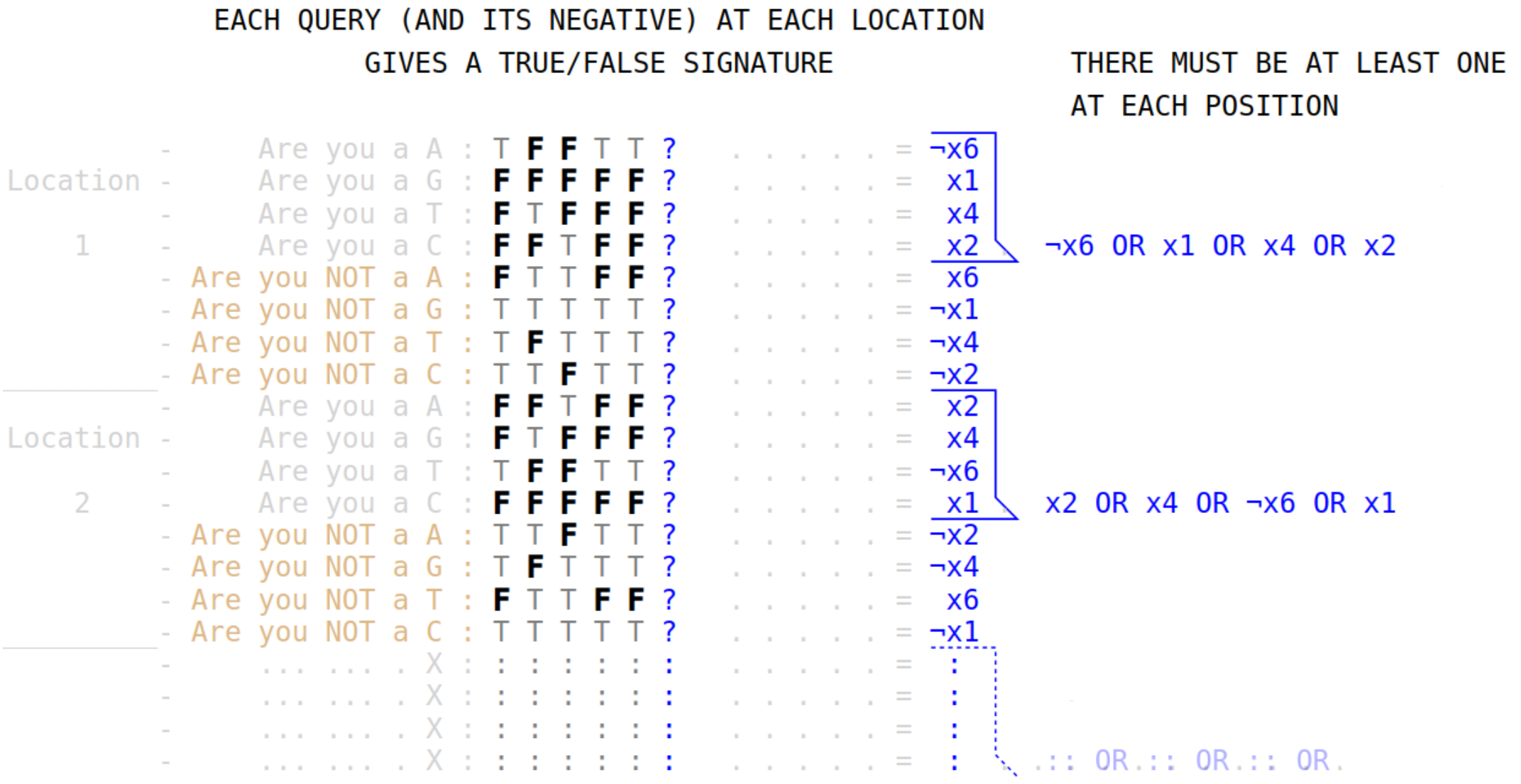

Supplemental Figure 1e: Queries and signatures across genome locations, and the associated binary variables determining the synthetic response to those queries (shown in blue against '?' characters), at-least-one A/T/G/C must be present at a location leads to size-4 clauses being generated and added.

In **Supplemental Figure 1e**, we see the queries and signatures associated with the real dataset (such as also shown in **Supplemental Figure 1b**), and against these signatures we can see that the synthetic data response to these queries and signatures is given by respective binary variables – such as established by process depicted in **Supplemental Figure 1c**).

In this context, that the synthetic data has “an A at position 1” is determined by the value of literal ¬x6, that the synthetic data has “a G at position 1” is determined by value of literal x1, that the synthetic data has “a T at position 1” is determined by value of literal x4, and that the synthetic data has “a C at position 1” is determined by the value of literal x2. Naturally, since no input genome has more than one A/T/G/C at location 1, constraints will be added to ensure that no two of A/T/G/C will be indicated for the synthetic data at such a location 1 (see previous **Supplemental Section 1f**) - notably this includes the set of constraints: ‘x6 OR ¬x1’, ‘x6 OR ¬x4’, ‘x6 OR ¬x2’, ‘¬x1 OR ¬x4’, ‘¬x1 OR ¬x2’, ‘¬x4 OR ¬x2’. However this is not sufficient to ensure that there is at-least-one indicated A/T/G/C at location 1 (i.e. Not indicated none of A/T/G/C at location 1). This is straightforwardly given by the constraint ‘¬x6 OR x1 OR x4 OR x2’ which is true only if any one of ¬x6, x1, x4 x2 are true, or equivalently, is false only if all ¬x6, x1, x4 x2 are false. In this way the addition of constraint ‘¬x6 OR x1 OR x4 OR x2’ ensures that there will be at-least-one A/T/G/C at location 1, and the series of 2-clauses ensure that there be at-most-one A/T/G/C at location 1 – together these ensure that there is exactly-one A/T/G/C at location 1.

This logic is repeated for all genome locations, for location 1 the synthetic response variables are ¬x6, x1, x4 x2 and thus the constraint ‘¬x6 OR x1 OR x4 OR x2’ is added to ensure that at least one A/T/G/C is indicated at location 1. At location 2 the synthetic response variables are x2, x4, ¬x6 x1 and thus the constraint ‘x2 OR x4 OR ¬ x6 OR x1’ is added. At location 3 there may be different response variables, which will result in a new constraint added, and so on. These size-4 clauses simply ensure that the SAT solver resolves that there is at least one (i.e. Not zero of) {A,T,G,C} at every location.

The process of adding these size-4 constraints is considered in the pseudocode of the algorithm (**Supplemental Figure 2**), particularly on lines 10,15,16: On line 10 the set of sets of signatures corresponding to positive queries (eg. “are you an A/T/G/C”) per location are collected; On line 15 and 16, for each set a size-4 clause is added to the SAT solver – one clause per location – ensuring that at-least-one of the binary variables associated with these positive queries (per location) must be true.

These constraints (as considered in this subsection and the previous) are input into a SAT solver, which then is sufficient for the SAT solver to resolve the values of all binary variables, which indicates the nucleotide information of the synthetic genome.

## Section 1h: SAT solving and genome resolving

The constraints of size-4 and size-2 (as considered in the last two subsections) are input into a SAT solver, which then solves for the values of all binary variables to satisfy those constraints, the values of those binary variables indicate the synthetic response to the respective queries, which in turn determine the nucleotides present at each genome location.

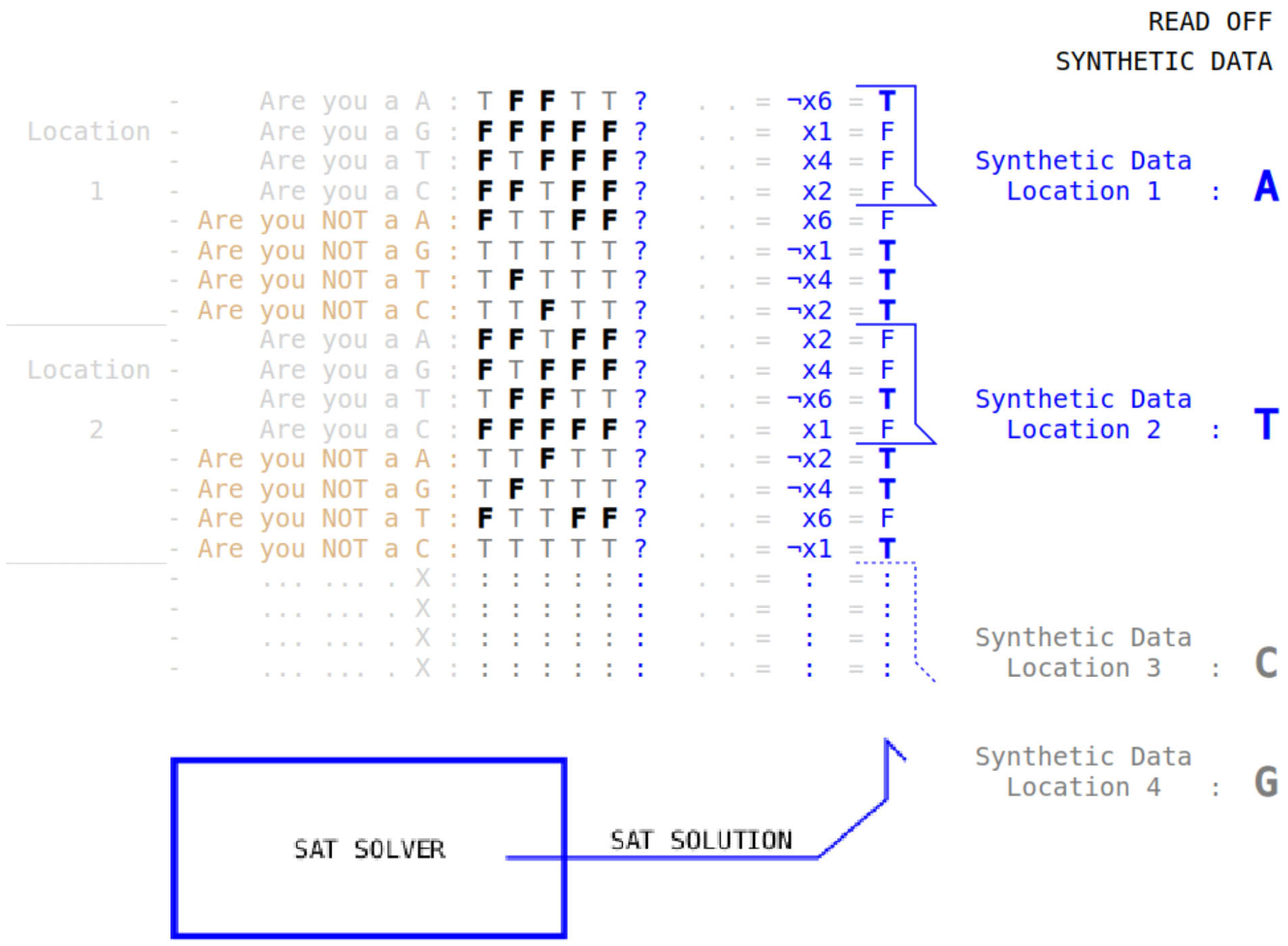

Supplemental Figure 1f: Queries and signatures across genome locations, and the associated binary variables determining the synthetic response to those queries (shown in blue against '?' characters) the values of all binary variables are determined by the solution to the SAT solver, from which the correct nucleotides of the synthetic data can then be directly read off.

This process is depicted in **Supplemental Figure 1f**, where we can see the synthetic response to query "are you an A at location 1" is given by the value of ¬x6, which the SAT solver evaluates to being 'true', and therefore the synthetic data is determined to have an 'A' nucleotide at location 1.

Conversely the synthetic response to alternative queries about location 1 - 'Are you a G/T/C at location 1?' given by variables x1,x4,and x2, which are all evaluated to be false, and the synthetic responses to opposite queries 'are you NOT an A/T/G/C at location 1?' are all evaluated oppositely to their positive counterparts, making this interpretation of their being an 'A' at location 1 unambiguous.

The values to all these binary variables are determined by the SAT solver consistent with all generated constraints (whose generation is considered in all previous subsections) and in this way the nucleotides of the entire synthetic genome are determined.

The resulting synthetic data illustrated (ATCGG) is then determined and can be compared with the input data shown in **Supplemental Figure 1a**. Where the primary rationale of the algorithm – stated in **Supplemental Section 1b** – can be inspected and seen to be satisfied.

The process of using a SAT solver to resolve these constraints and transform the values of the resulting binary variables back into genome data is considered in pseudocode (**Supplemental Figure 2**) on lines 21-27: This is where a SAT solver execution call is made on line 21, and an empty array is declared to be filled with synthetic genome data on line 22. The input into the algorithm is re-read per location on line 23, and for each nucleotide on line 24, the positive signature 'k' is calculated (for that nucleotide at that location), on line 25. This reflects the construction and consideration (in turn) of all positive signatures (corresponding to queries ‘Are you an A/T/G/C?’ across locations – in turn) such as depicted in **Supplemental Figure 1f**. Where the binary variable associated with such a positive signature is ‘True’ - on line 26, the corresponding nucleotide A/T/G/C is sequentially added to the array. Which is the output of the algorithm on line 28.

## Section 1i: Implementation details and remarks

The software implementation of Genomator takes an input of genome samples in a VCF file. However VCF files do not necessarily indicate single nucleotides across genome positions but may indicate the presence of one or many variants (reference and alternative/s) and between different genome positions and also chromosomes phased or unphased.
In the explanation of the algorithm considered thus far, we have considered single nucleotides for ease of illustration. However, the full algorithm handles this complexity by considering more specific queries that incorporate location and phase in the construction of synthetic data.
With single nucleotides, more simple queries are considered – eg. “Are you an ‘A’ at location 1?”, whereas for more complex genome data, more complex queries can be considered and folded into the algorithm – eg. “do you have alternative structural feature #1 on phase A at location X?” etc.
In this way the full feature set of VCF files can be generated using Genomator; we note that diploid and haploid datasets are considered, generated and evaluated, as documented in the main paper.

As considered in in **Supplemental Section 1f**, Genomator considers all pairs of signatures in the construction of size-2 constraints. However there is nothing prohibiting the consideration of triplets (consequently generating size-3 constraints) or quadruplets of signatures (consequently generating size-4 constraints).
Were Genomator constructed to consider all quadruplets of signatures, then it would accomplish eliminating from the synthetic data all quadruplets of features not present in the input data - which is a significantly more constricting requirement.
However size-2 clauses are particularly amenable for solving using backtracking SAT solvers (noting that a SAT problem entirely composed of size-2 constraints – known as 2SAT is demonstrably polynomially time soluble as opposed to the more general case which is known to be NP-complete), and there are significantly fewer pairs of variants in genome data sets than there are combinations of quadruplets for the software to handle.

As considered in **Supplemental Section 1c**, the larger the input dataset the looser the constraints become in the construction of the synthetic data, and where the input dataset is quite sizeable potentially becomes too loose to reasonably reconstruct sensible output.
This issue can be resolved several ways, however the principle means used in the software to resolve this issue is to create clusters of similar inputs (of size N, where similarity is by hamming distance) and to randomly and iteratively input these clusters into the Genomator algorithm to output a range of synthetic data.
The clustering of input data is done prior to execution, and multiple overlapping clusters are randomly generated, such that no input sample is under-represented across those input clusters (to avoid biasing).
As stated in **Supplemental Section 1c**, the larger the size of the cluster inputs into genomator (size N) and the more varied the genomes in those clusters are, the less constrained (therefore potentially more random) Genomator's output has the potential to be.

In practice the larger the value of N, the looser the constraints that the SAT solver has to solve, conversely the larger the value of Z, the more rarer pairs of SNPs are attenuated and the more constricting the constraints that the SAT solver must solve.
If the constraints are too restricting the SAT solver may be unable to solve the solution (either demonstrating that there is no possible solution, or timing-out), thus for higher values of Z, the greater the value of N is necessary for the synthetic data output to be possible.

We note that the cluster size N and stochastic parameter Z controls how much randomisation occurs in the construction of the constraints that Genomator generates, and these are core parameters that affects the accuracy and privacy afforded by the process. (NOTE: where in the text we indicate that Z has a particular value α, we are indicating that the algorithm draws from a uniform distribution on [0, α) and is then rounded down to integer value)

In our depiction and explanation of Genomator's mechanism we have focused on the process of generating a single synthetic output, however in practice we generate multiple outputs by re-running the process. Rerunning Genomator on the same input will likely yield different results (because SAT solvers can possess some degree of randomness in generating solutions within constraints) but we also re-run with different sets of inputs by randomly inputting different input clusters from. These factors add to give randomisation to Genomator's output. However, and optionally, additional constraints can be added to the SAT solver to ensure that different outputs are unique from each other and/or also the inputs, by some minimum hamming distance. We call these diversity constraints, and are an option provided by the software – but not employed in the results of our paper.

The PySAT library[1] was used to provide the SAT solver interface in the software package. PySAT provides a common interface to a range of different SAT solvers which have different functionality and strategies, notably we employed the MiniCard[3] SAT solver which is a simple extension of the well known MiniSat[4] solver built with the ability to also handle cardinality constraints natively. SAT solvers of this sort are called 'conflict driven clause learning' (CDCL) solvers (a subset of backtracking SAT solvers) and proceed by assigning values to binary variables in the problem to systematically solve a set of constraints. Notably, backtracking SAT solvers are particularly efficient at solving SAT problems with size-2 clauses - which are the majority of the constraints used in Genomator's algorithm.

The algorithm involves conversion of genome data into binary signatures, and reasons over pairwise combinations of unique signatures in order to ensure that all pairs of features that are not in the input are not reproduced in the synthetic data. The simplification of assigning unique binary variables only to unique signatures (the deduplication described in **Supplemental Section 1e**) provides a significant simplification of the problem, and as shown in the main paper, Genomator can generate synthetic genomes of 11M variants in a number of seconds, ensuring that all possible features pairs not occurring across 11M variants are not reproduced. This performance is created by this simplification, as there are at-most only 2^10 unique signatures (therefore 2^9=512 binary variables) ising from an input cluster size of N=10.

We note that this simplification is, in principle, optional and the central rationale of Genomator's algorithm (per **Supplemental Section 1b**) could be satisfied by using a SAT solver and assigning a unique binary variables to each of the queries about the synthetic data (such as shown in **Supplemental Section 1b**). And in that case 11M variants would result in ~11M binary variables instead of 2^9=512.

The simplification that inverted signatures are associated with inverted binary variables also simplifies the SAT problem.

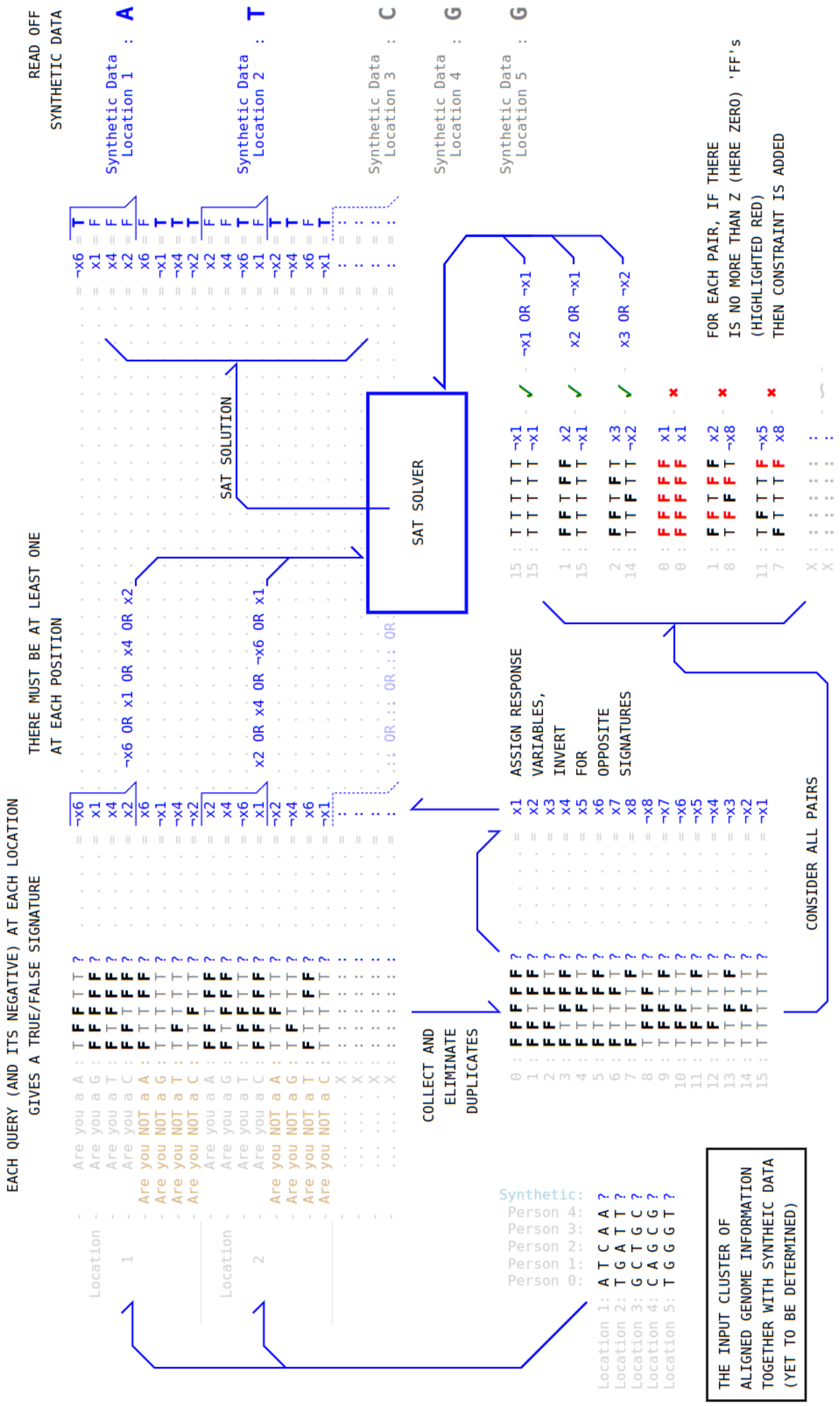

Supplemental Figure 1: Flowchart of the processing steps used by Genomator in creation of synthetic genomic data. The process is input with a cluster of aligned genome data and then computes binary vectors from these to a set of logical conditions (eg. "Are you A,T,G,C?" and their inverses) these vectors are then deduplicated and considered pairwise in the construction of constraints for a SAT problem. The solutions to the SAT problem are then reinterpreted as synthetic genomic information. For illustration purposes we show single nucleotides across genome locations, but the algorithm can extend to process diploid genomes. Undetermined information is denoted by blue "?" characters, which are then set to be determined by the values of specific binary variables (also in blue).

```
function GENOMATOR(genomes,N,𝒵 ∈ ℕ)
    ▷ genomes_{m,n} is dataset neucleotide at position m, of sample n ◁
    select_genomes_{m,n} ← SELECTNSAMPLES(genomes,N)
    unique_{m,n} ← SORT({select_genomes_{j,·}}_j)    ▷ Deduplicating by positions
    signatures ← ∅, clauses ← ∅
    for all v ∈ [A, T, G, C] do    ▷ Constructing all unique binary 'signatures'
        Q_{j,i} ← unique_{j,i} = v
        W_{j,i} ← unique_{j,i} ≠ v
        signatures ← signatures ∪ {Q_{j,·}}_j ∪ {W_{j,·}}_j
        clauses ← clauses ∪ {{Q_{j,·}}_j}
    n ← SIZE(signatures)/2
    signatures ← SORTBYINDICES(signatures)
    variables ← [x_1, ..., x_n, ¬x_n, ..., ¬x_1]
    f(s) = variables[at INDEX(s)], for s ∈ signatures
    for all c ∈ clauses do    ▷ Adding 'Reverse Readability' AtLeastOnes
        ADD_CLAUSE(sat_solver,{f(s) | s ∈ c})
    for all s1,s2 ∈ signatures do    ▷ Adding 'Indistinguishability' constraints
        DRAW z ∼ 𝒵    ▷ Stochastic strengthening
        if less than z false of {s1_i ∨ s2_i}_i then
            ADD_CLAUSE(sat_solver,{f(s1), f(s2)})
    solution ← SOLVE(sat_solver)
    synth ← EMPTYARRAY
    for all t ∈ {select_genomes_{·,j}}_j do    ▷ Transforming into Neucleotides
        for all v ∈ [A, T, G, C] do
            k_i ← t_i = v
            if f(k) ∈ solution then
                APPEND(synth,v)
    return synth
```

Supplemental Figure 2: Pseudocode for the Genomator algorithm.

## Supplemental Section 2: Detailed description of Reverse Genomator

As previously discussed, Genomator employs logical deduction to create a range of synthetic data output from a subset of input genomes. This means that the deduction process can in-principle be reversed: from a synthetic data output, the range of possible sets of input genomes which could have been used to generate it can be inferred. This functionality is executed by our auxiliary tool, Reverse Genomator, which provides a means assessing the privacy afforded by Genomator.

Reverse Genomator constructs a SAT problem which yields plausible subsets of genomes, which could have given rise to a witnessed synthetic output and therefore could plausibly have been the subset that was actually used in such generation. Between these subsets, the preponderance of presence of individual genomes gives an approximation of the *a posteriori* likelihood that those genomes were actually used in the generation of the witnessed output

- as described in **Supplemental Figure 6**. And most particularly, if a specific input appears in all the sets resolvable by Reverse Genomator then it is necessarily in the input – assuming that the superset of plausible genomes is correctly known with confidence.

Privacy, considered and measured in this way (as depicted in **Supplemental Figure 3**) approximates the Bayesian confidence that an attacker could have about which individuals were used in generating a witnessed output under an extreme scenario where the attacker has access to all the original dataset sequences and only possesses ignorance about which subset was used in the generation of specific synthetic output – see **Supplemental Figure 3**. In this way Reverse Genomator serves to provide a proxy for the absolute privacy provided by Genomator by deductive reverse inspection of Genomator's output.
We note that this attack model is known as a 'membership inference attack' - where an attacker seeks to know if a particular input was part of the input dataset - which contrasts against the 'attribute inference attack' as considered in Section 3.3 - where an attacker seeks to infer specific information about a target under assumption that the individual was part of the input dataset.

Within the flowchart of the process (in **Supplemental Figure 4**, accompanying pseudocode **Supplemental Figure 5**), the witnessed synthetic data is considered alongside the aligned genomic information from all the potential input genomes, and a SAT problem is constructed with binary variables indicating the inclusion of each potential input in a candidate set. Reverse Genomator then formulates SAT constraints such that wherever there is a pair of logical conditions which the synthetic genome satisfies, then at least stochastic parameter 'Z' of the input genomes which also satisfies that pair of conditions, must also be included in any candidate set.

Enumeration over all pairs of logical conditions yields a large number of SAT constraints, many of which are redundant by subsumption. The remaining SAT constraints are then passed into a SAT solver which then resolves a candidate set of genome inputs which could have been used to generate the witnessed output. Multiple solutions can then be generated from the SAT solver, which indicates a range of candidate input sets. Between these candidate sets the preponderance of specific individual genomes in them provides a proxy for the likelihood that those genomes were in the input dataset that actually was used to generate the witnessed output.

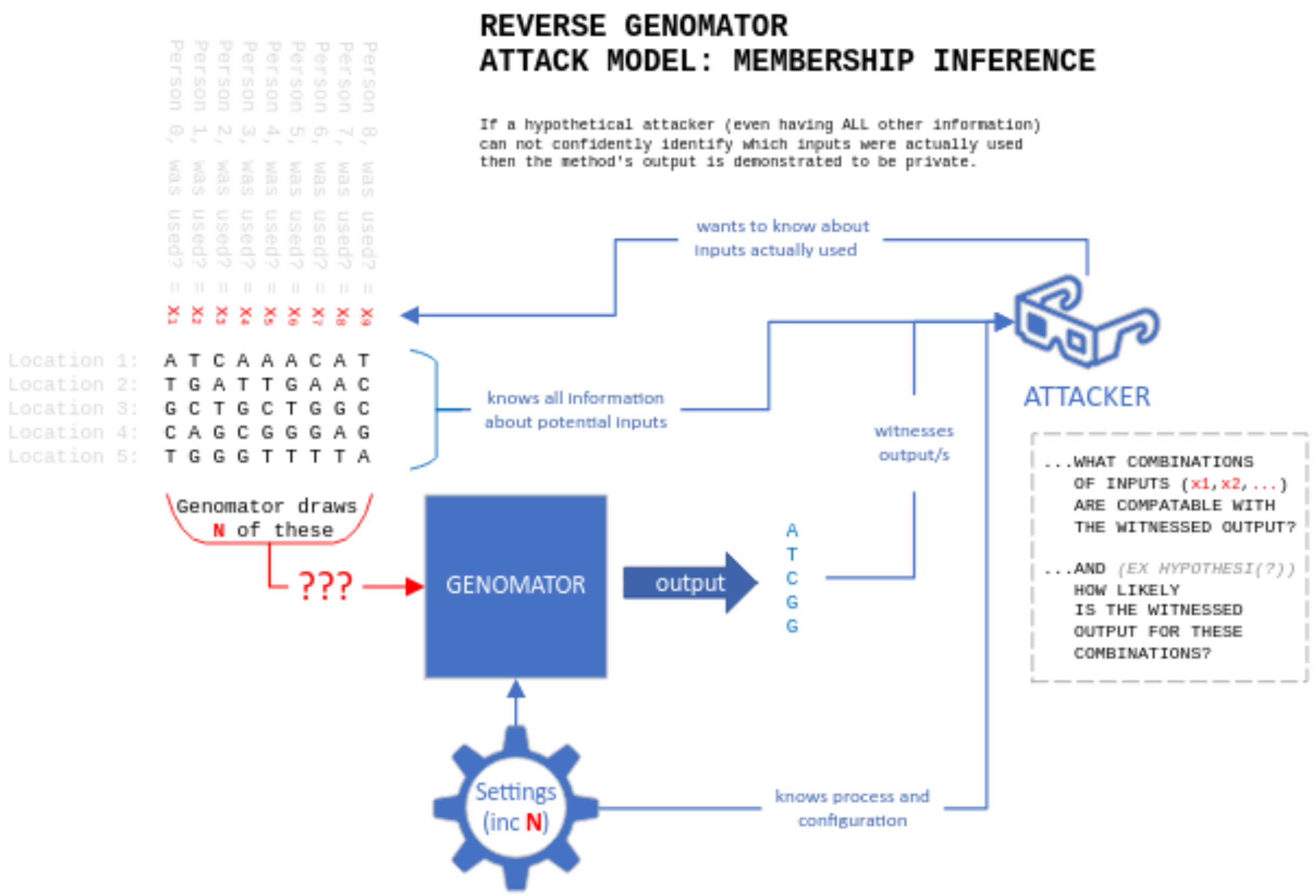

Supplemental Figure 3: Illustration of the attack model in which Reverse Genomator is applicable, particularly when a hypothetical attacker wishes to know membership of inputs actually used from a superset of plausible genomes. For illustration purposes we show single nucleotides, but the algorithm can process diploid genomes.

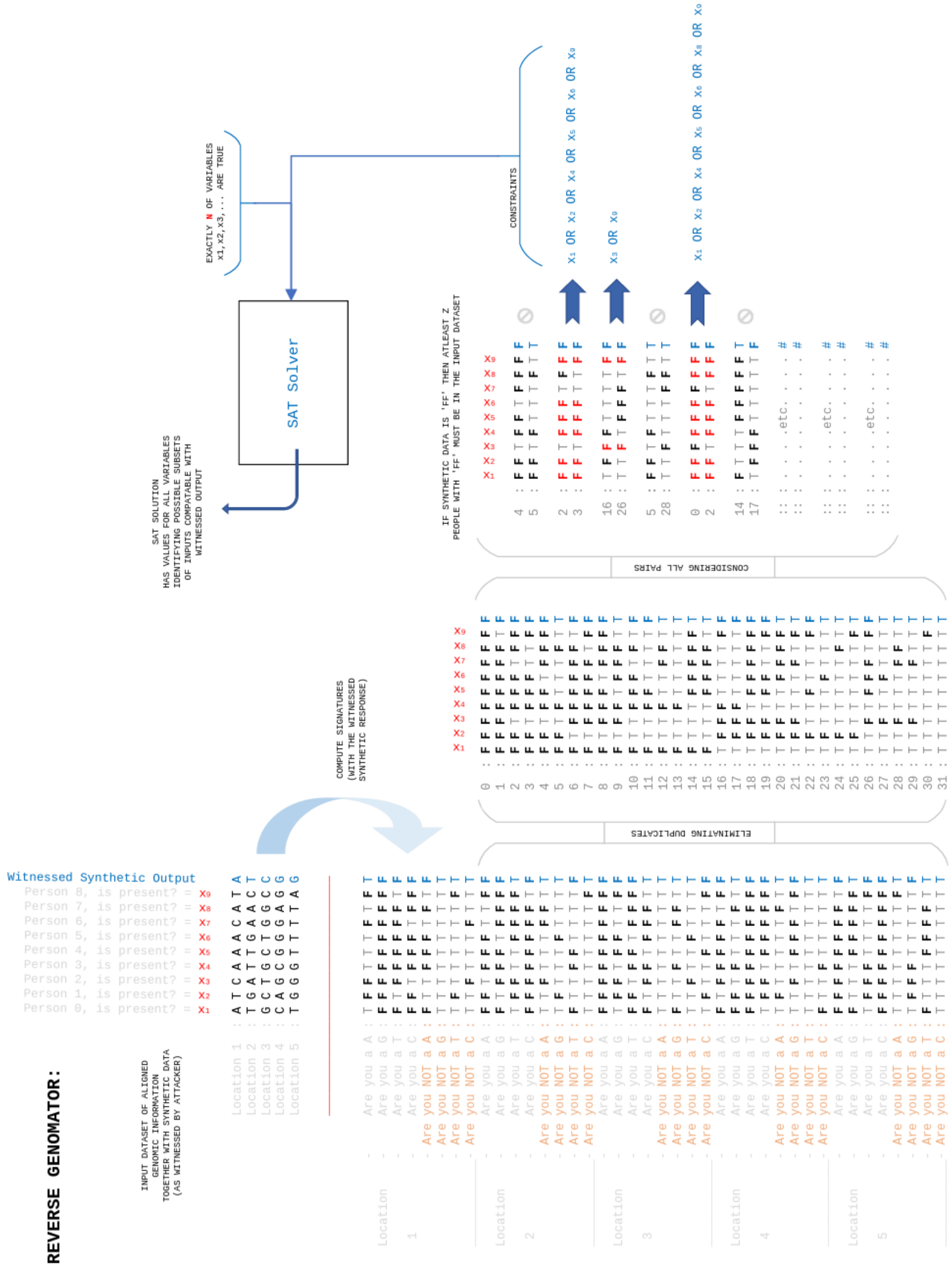

Supplemental Figure 4. Flowchart of the processing steps used by Reverse Genomator in the analysis of synthetic data produced by Genomator. The process considers possible input genomes which could have been used to generate the witnessed output from Genomator indicated by binary variables, and a SAT problem is constructed to resolve possible sets which would be compatible with the generation of the witnessed output. For illustration purposes we show single nucleotides but the algorithm can process diploid genomes.

**function** REVERSE_GENOMATOR(synth,genomes,N,trials,$\mathcal{Z} \in \mathbb{N}$)

  ▷ *$synth_m$ is synthetic neucleotide at position $m$* ◁

  ▷ *$genomes_{m,n}$ is neucleotide at position $m$, of sample $n$* ◁

  signatures $\leftarrow \emptyset$

  **for all** $v \in [A, T, G, C]$ **do** ▷ *Extracting relevent binary 'signatures'*

    $\mathrm{Q}_{j,i} \leftarrow \mathrm{genomes}_{j,i} = v$

    $\mathrm{W}_{j,i} \leftarrow \mathrm{genomes}_{j,i} \neq v$

    signatures $\leftarrow$ signatures $\cup \{\mathrm{Q}_{j,\cdot} \mid \mathrm{synth}_j \neq v\} \cup \{\mathrm{W}_{j,\cdot} \mid \mathrm{synth}_j = v\}$

  half_clauses $\leftarrow \{\{x_i \mid \neg s_i\}\}_{s \in \mathrm{signatures}}$

  solutions $\leftarrow$ EMPTYARRAY

  **for** $k \in [1, \ldots, \mathrm{trials}]$ **do** ▷ *Solving $k$ times SAT problems*

    CLEAR(sat_solver) ▷ *Solved $x_i$ indicate possible* GENOMATOR *input*

    L $\leftarrow \{$LESSTHANZFALSEOF($c_1 \cap c_2$, DRAW $z \sim \mathcal{Z}$)$\}_{c_1, c_2 \in \mathrm{half_clauses}}$

    ADDCLAUSES(sat_solver, ELIMINATESUBSUMEDCONSTRAINTS(L))

    ADDEXACTLYNTRUECONSTRAINT(sat_solver, $\{x_i\}_i$, N)

    APPEND(solutions, SOLVE(sat_solver))

  **return** $\{$COUNT$(\{x_i \in \mathrm{solutions}_j\}_j) \div \mathrm{trials}\}_i$

Supplemental Figure 5: Pseudocode for the Reverse Genomator algorithm.

**Supplemental Theorem 1:**

**Theorem 1.** *In the combinations $C_{1\ldots q}$ of inputs provided by Reverse Genomator that are compatible with the witnessing $O$ of output $o$ from Genomator, the product of the ratio of those combinations that do not feature the target individual $i$ over those that do ($\zeta$), and the average inverse model count ratio ($R$), gives the confidence an attacker should have for the inference $I$ that individual $i$ was part of the dataset that yielded $o$, as:*

$$P(I|O) = \frac{1}{1+\zeta \cdot R}$$

*where* $\zeta = \frac{N_{\notin}}{N_{\in}}$ *and* $R = \frac{\mathbb{E}\left[\frac{1}{|f(C_k)|}|i \notin C_k\right]}{\mathbb{E}\left[\frac{1}{|f(C_k)|}|i \in C_k\right]}$

*Here, $N_{\notin}$ is the number of compatible combinations not containing $i$, $N_{\in}$ is the number of compatible combinations containing $i$, $\mathbb{E}[\cdot]$ denotes expectation, and $f(C_k)$ is the set of possible outputs from Genomator for a given input combination $C_k$.*

**Assumption 1.** *Assuming the SAT solver is unbiased such that no output is more likely than any other, for any combination $C_k$ of inputs provided by Reverse Genomator that are compatible with the witnessing $O$ of output $o$:*

$$P(O|C_k) = \frac{1}{|f(C_k)|}$$

*where $|f(C_k)|$ is the total number of possible outputs for the input combination $C_k$.*

*Proof.* We consider Bayes' theorem:

$$P(I|O) = \frac{P(O|I)P(I)}{P(O|I)P(I) + P(O|\neg I)P(\neg I)}$$

$$P(I|O) = \frac{1}{1 + \frac{P(O|\neg I)P(\neg I)}{P(O|I)P(I)}}$$

Using the law of total probability over compatible combinations $C_k$:

$$\frac{P(O|\neg I)P(\neg I)}{P(O|I)P(I)} = \frac{\sum_k P(O|C_k)P(C_k|\neg I)P(\neg I)}{\sum_k P(O|C_k)P(C_k|I)P(I)}$$

Applying Bayes' theorem to $P(C_k|\neg I)$ and $P(C_k|I)$:

$$= \frac{\sum_k P(O|C_k)P(\neg I|C_k)P(C_k)}{\sum_k P(O|C_k)P(I|C_k)P(C_k)}$$

Using our assumption 1:

$$= \frac{\sum_{k:i\notin C_k} \frac{1}{|f(C_k)|}P(C_k)}{\sum_{k:i\in C_k} \frac{1}{|f(C_k)|}P(C_k)}$$

If each $C_k$ is equally likely to be selected:

$$= \frac{\sum_{k:i\notin C_k} \frac{1}{|f(C_k)|}}{\sum_{k:i\in C_k} \frac{1}{|f(C_k)|}}$$

Leading to:

$$\frac{P(O|\neg I)P(\neg I)}{P(O|I)P(I)} = \underbrace{\frac{N_{\notin}}{N_{\in}}}_{\text{Quantity Ratio, } \zeta} \cdot \underbrace{\frac{\mathbb{E}\left[\frac{1}{|f(C_k)|} \mid i \notin C_k\right]}{\mathbb{E}\left[\frac{1}{|f(C_k)|} \mid i \in C_k\right]}}_{\text{Inverse Model Count Ratio, } R}$$

Therefore:

$$P(I|O) = \frac{1}{1+\zeta \cdot R}$$

$\zeta$, is a simple count ratio, and $R$ captures how the presence or absence of $i$ affects the average number of possible outputs. When $R$ is close to 1, $\zeta$ becomes the primary factor in determining an attacker's confidence about whether $i$ was part of the dataset. □

Supplemental Figure 6: Theorem associating the preponderance of counts in Reverse Genomator to the confidence that a hypothetical attacker can have about inputs.

## Supplemental Section 3: Overview of other methods used in the benchmark.

**Markov Chains:** are stochastic models of system dynamics where the next state of a system depends on the previous state. In the context of synthetic genome generation, these algorithms statistically construct the next nucleotide based on the previously generated nucleotides in a scan across nucleotide positions. A primary implementation and analysis of this kind of generation method is given by Samani *et al*[5]. In this context, the next nucleotides are determined by conditional probabilities as measured over the source dataset, and a primary parameter of this method is how many previously generated nucleotides are considered in generating the next, this is called the 'window size'.

Between different datasets the width of the window is an important parameter to choose. This choice is done to capture the relevant information between important SNPs without making the window too large such that tracts of individual specific information get reproduced. Code for this method was adapted from that used in Yelmen *et al.* 2021[6], with repository https://gitlab.inria.fr/ml_genetics/public/artificial_genomes.

**Generative Adversarial Networks:** (GANs) are a bi-modal neural network composed of a generator and a discriminator model[7]. The generator model produces synthetic genomic sequences while the discriminator is trained to detect whether the sequences generated are real or synthetic. Based on the feedback provided by the discriminator model, the generator model modifies the network's weights to make the sequences even more indistinguishable, which in-turn then serves to refine the discriminator network.

Code for this method was from that used in Yelmen *et al.* 2021[6], with repository https://gitlab.inria.fr/ml_genetics/public/artificial_genomes (retrieved March 2024). The generator architecture was a series of fully connected layers, one input layer to 2 hidden layers (size a factor of 1.2 and 1.1 smaller than the dimension of the input) and an output layer. For the discriminator is a smaller model with 2 hidden layers of size a factor of 2 and 3 smaller than the dimension of the input. Both models were trained over the suggested 20000 epochs. The implementation used ADAM[8] as optimization algorithm with cross entropy loss function. To investigate the variability provided by GANs method we added an optional multiplier on the size of all these neural layers.

Please note that the newest configuration of GANs by Yelmen *et al.* 2023[9], Wasserstein GANs (WGAN), could not be used in our paper. The architecture of their provided implementations was tailored to the specific SNP numbers of 10K and 65K (see https://gitlab.inria.fr/ml_genetics/public/artificial_genomes/-/tree/master/WGAN) used in their paper, which – unlike the previous version (GANs) – was not transferable to our datasets: We first received an RuntimeError "Trying to create tensor with negative dimension -1: [1, -1]" which was due to

latent_depth_factor = 14 #14 for 65535 SNP data and 12 for 16383 zero padded SNP data

We successfully got past this point with latent_depth_factor=4, however received a new RuntimeError "Sizes of tensors must match except in dimension 1. Expected size 204799 but got size 805 for tensor number 1 in the list."

We concluded that an extensive re-architecting would be needed to apply WGANs to new datasets. Since, the reported difference between WGANs and GANs was a correlation coefficient improvement from 0.94 to 0.96, we opted for using the more generalizable GAN code especially since GANs reached in our hands similar high performances, see **Supplemental Table 3.**

**Restricted Boltzmann machine:** (RBMs) are a type of generative neural network model that can learn the underlying probability distributions over a set of input data. They consist of two layers of units: the visible layer which represents the input data and the hidden layer which represents the features/patterns of the input data. The units in each layer are fully connected by weights, which measures how strongly they influence each other. The energy of a joint configuration of visual and hidden units is determined by the sum of activated weights and biases, where the lower the energy the more likely the configuration of the network is. The training process consists of updating these weights via a sampling process with gradient descent (called contrastive divergence), which compares the input data with the generated data and tries to reduce the difference between them. After training, RBMs can be used to generate new data consistent with the learned probability distribution. The RBM model and code for this method was adapted from that used in paper Yelmen *et al.* 2021[6], with repository https://gitlab.inria.fr/ml_genetics/public/artificial_genomes, and was implemented using python using PyTorch library version 1.10.2, where the RBM model consisted of 500 nodes in the hidden layer. The RBM method in our hands perform comparably to what is reported by Yelmen *et al.* 2023[6] see **Supplemental Table 3.**

**Conditional Restricted Boltzmann machine:** Conditional Restricted Boltzmann machines (CRBMs) are an adaptation of RBM architecture considered by Yelmen *et al.* 2023[9] as a means of more effectively scaling RBM technique to larger genome data. The process consists of training an RBM on an initial section of the training data, and then additional RBMs on consequent sections of data conditional on section of previous data. The CRBM model and code for this method was adapted from that used in their paper as found in https://gitlab.inria.fr/ml_genetics/public/artificial_genomes and was implemented using python using PyTorch library version 1.10.2.

We found that the defaults in their code repository did not on our data yield the performance reported in their paper (**Supplemental Figure 7**). We hence used the CRBM model with 500 nodes in the hidden layer and a fixed-node overlap of 300, which represents the input data more closely (**Supplemental Figure 8**).

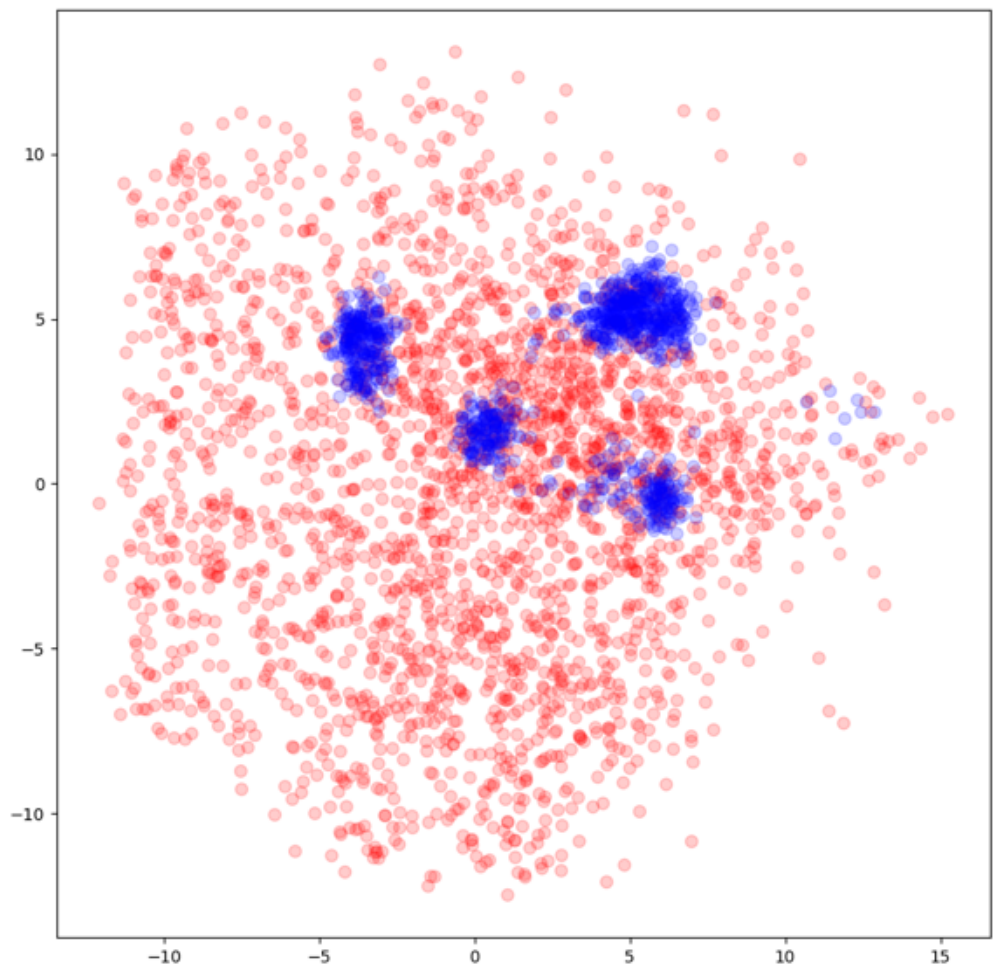

Supplemental Figure 7: PCA of generated synthetic data produced by CRBM method (blue) against input dataset (red) of the AGBL4 gene, produced with 2 days of runtime, 900 epochs, 2000 hidden nodes in hidden layer, 5000 fixed nodes per model, learning rate of 0.001, minibatch size of 1252, and 1252 parallel chains to compute the negative term of the gradient in training.

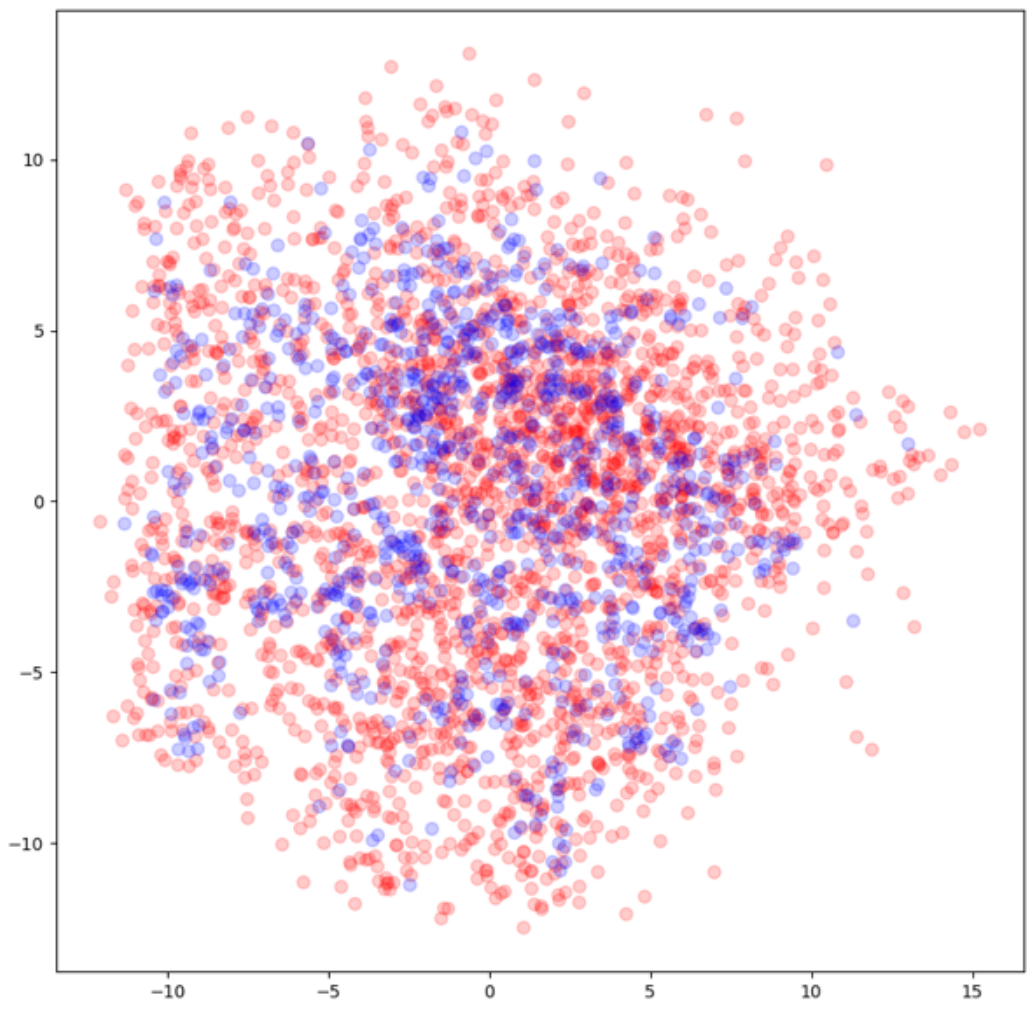

Supplemental Figure 8: PCA of generated synthetic data produced by CRBM method (blue) against input dataset (red) of the AGBL4 gene, produced with ~30 minutes of runtime, 1200 epochs, 500 hidden nodes in hidden layer, 300 fixed nodes per model, learning rate of 0.005, minibatch size of 500, and 500 parallel chains to compute the negative term of the gradient in training.

## Supplemental Section 4: Comparison of methods' stabilities

We show the variability of the 5 methods we evaluate by calculating the Wasserstein distance between synthetic genomes and source dataset PCAs for multiple runs.

| Run Number | Wasserstein Distance | Standard Deviation |
|---|---|---|
| Run1 | 1643.1 | 16.6 |
| Run2 | 1829.2 | 18.8 |
| Run3 | 1644.6 | 9.1 |
| Run4 | 2649.4 | 58.4 |
| Run5 | 2707.7 | 15.2 |
| Run6 | 2092.2 | 17.4 |
| Run7 | 2471.3 | 59.6 |
| Run8 | 1312.4 | 9.4 |
| Run9 | 1735.7 | 22.4 |
| Av (std) | 2009.5 (497.2) | |

Supplemental Table 1a: Wasserstein distance between real and synthetic data generated by GAN method on 805 SNP data.

| Run Number | Wasserstein Distance | Standard Deviation |
|---|---|---|
| Run1 | 911.5 | 12.4 |
| Run2 | 790.7 | 13.0 |
| Run3 | 1976.2 | 46.0 |
| Run4 | 917.8 | 18.8 |
| Run5 | 1648.4 | 37.0 |
| Run6 | 1621.3 | 22.4 |
| Run7 | 1106.5 | 17.8 |
| Run8 | 2104.4 | 54.6 |
| Run9 | 2571.9 | 61.7 |
| Av (std) | 1516.5 (624.3) | |

Supplemental Table 1b: Wasserstein distance between real and synthetic data generated by RBM method on 805 SNP data.

| Run Number | Wasserstein Distance | Standard Deviation |
|---|---|---|
| Run1 | 7230.8 | 66.4 |
| Run2 | 7265.5 | 65.0 |
| Run3 | 7301.9 | 67.9 |
| Run4 | 7208.5 | 64.7 |
| Run5 | 7237.7 | 65.3 |
| Run6 | 7285.9 | 71.1 |
| Run7 | 7154.9 | 60.7 |
| Run8 | 7329.3 | 70.3 |
| Run9 | 7225.7 | 68.7 |
| Av (std) | 7248.9 (52.9) | |

Supplemental Table 1c: Wasserstein distance between real and synthetic data generated by Markov method on 805 SNP data.

| Run Number | Wasserstein Distance | Standard Deviation |
|---|---|---|
| Run1 | 1020.4 | 5.8 |
| Run2 | 1534.4 | 26.4 |
| Run3 | 1145.1 | 12.9 |
| Run4 | 1057.9 | 14.3 |
| Run5 | 1878.5 | 33.2 |
| Run6 | 1062.6 | 13.1 |
| Run7 | 1531.8 | 9.6 |
| Run8 | 1417.9 | 29.8 |
| Run9 | 1786.0 | 36.3 |
| Av (std) | 1381.6 (326.0) | |

Supplemental Table 1d: Wasserstein distance between real and synthetic data generated by Genomator method on 805 SNP data.

| Run Number | Wasserstein Distance | Standard Deviation |
|---|---|---|
| Run1 | 9641.4 | 140.4 |
| Run2 | 9707.7 | 162.2 |
| Run3 | 9757.5 | 160.2 |
| Run4 | 9666.3 | 161.5 |
| Run5 | 9900.1 | 154.2 |
| Run6 | 9872.4 | 143.9 |
| Run7 | 9721.6 | 145.6 |
| Run8 | 9611.3 | 140.3 |
| Run9 | 9876.5 | 162.0 |
| Av (std) | 9750.5 (108.5) | |

Supplemental Table 1e: Wasserstein distance between real and synthetic data generated by CRBM method on 805 SNP data.

## Supplemental Section 5: LD Accuracy

For each gene, including introns, exons, and untranslated regions, pre-processing was performed on genetic variants from 1KGP project using PLINK2[10] as detailed in our paper. In the main text, we presented figures illustrating the LD square error of synthetic data at incrementing windows (Figure 2c). Additionally, we calculated the average LD and average square error in LD reproduction across all pairwise relationships in each dataset.
The pairwise LD values (and square error in LD reproduction values) were binned according to the difference in their index. The average LD square error within each bin was computed, followed by an overall average across all bins. This method was employed to give approximately equal weight to LD reproduction between loci of varying distances.
Pairwise LD consistency between synthetic and original dataset was calculated using Rogers-Huff r-squared method for LD[11] and shown in table **Supplemental Table 2a**. The square error as percentage of the observed Av LD is shown for each method in **Supplemental Table 2b**.
For consistence with the values reported by Yelmen et al.[9] we also provide the correlation coefficient in **Supplemental Table 3**.

| Gene | SNPs | Av. LD | Genomator | Markov | GAN | RBM | CRBM |
|---|---|---|---|---|---|---|---|
| CCSER1 | 7447 | 0.00935 | 0.000181 | 0.00199 | 0.00370 | 0.00125 | 0.00164 |
| AGBL4 | 3118 | 0.02090 | 0.000437 | 0.00903 | 0.00184 | 0.00317 | 0.00631 |
| FHIT | 8075 | 0.00527 | 0.0001117 | 0.000662 | 0.000857 | 0.000515 | 0.000617 |
| RBFOX1 | 14382 | 0.00523 | 0.0000969 | 0.000482 | 0.000599 | 0.000379 | 0.000471 |
| | | Average (std) | 0.000206 (0.000157) | 0.00304 (0.00404) | 0.00174 (0.00140) | 0.00132 (0.00128) | 0.00225 (0.00274) |

Supplemental Table 2a: average square error in reproducing LD r2, for different methods across several genes

| Gene | SNPs | Av. LD | Genomator | Markov | GAN | RBM | CRBM |
|---|---|---|---|---|---|---|---|
| CCSER1 | 7447 | 0.00935 | 1.94% | 21.28% | 39.57% | 13.37% | 17.54% |
| AGBL4 | 3118 | 0.0209 | 2.09% | 43.21% | 8.80% | 15.17% | 30.19% |
| FHIT | 8075 | 0.00527 | 2.12% | 12.56% | 16.26% | 9.77% | 11.71% |
| RBFOX1 | 14382 | 0.00523 | 1.85% | 9.22% | 11.45% | 7.25% | 9.01% |
| | | Average (std) | 2.00% (0.13) | 21.57% (15.30) | 19.02% (14.04) | 11.39% (3.56) | 17.11% (9.42) |

Supplemental Table 2b: percentage of average square error in reproducing LD r2 values relative to average LD, for different methods across several genes

| Gene | SNPs | Genomator | Markov | GAN | GAN_p2 | RBM | CRBM |
|---|---|---|---|---|---|---|---|
| CCSER1 | 7447 | 0.99736 | 0.99903 | 0.40674 | 0.98676 | 0.99827 | 0.94422 |
| AGBL4 | 3118 | 0.99862 | 0.99933 | 0.97013 | 0.93473 | 0.96938 | 0.89496 |
| FHIT | 8075 | 0.99754 | 0.99908 | 0.3738 | 0.98989 | 0.99782 | 0.96315 |
| RBFOX1 | 14382 | 0.99749 | 0.99917 | 0.45963 | 0.98996 | 0.99512 | 0.96645 |

Supplemental Table 3: Correlation coefficient (cc) of allele frequencies between synthetic and source datasets, between different methods and on different gene datasets. GAN denotes the cc resulting from using parameter settings proposed by Yelmen, while GAN_p2 denotes parameters tuned for this dataset (latent size of 400, generator and discriminator learning rate of 0.0004 and 0.0016, layer sizes of 800x800 and 600x400, trained for 300 epochs).

## Supplemental Section 6: Dataset Gene Information

| Name | Description | Position | RefSeq origin |
|---|---|---|---|
| AGBL4 | Homo sapiens ATP/GTP binding protein like 4 | hg38 chr1:48,532,854-50,023,954 | NR_136623 |
| FHIT | Homo sapiens fragile histidine triad diadenosine triphosphatase | hg38 chr3:59,747,277-61,251,452 | NM_002012 |
| CCSER1 | Homo sapiens coiled-coil serine rich protein 1 | hg38 chr4:90,127,394-91,605,295 | NM_001377987 |
| RBFOX1 | Homo sapiens RNA binding fox-1 homolog 1 | hg38 chr16:6,019,024-7,713,340 | NM_018723 |

Supplemental Table 4: Reference for Genes used in analysis

## Supplemental Section 7: Efficiency at scale across human genome

| SNPs | Genomator | Markov | GAN* | RBM* | CRBM* |
|---|---|---|---|---|---|
| 100 | 0.84 | 0.96 | 3610.32 | 7.54 | 9.27 |
| 400 | 0.94 | 2.52 | 3611.70 | 7.82 | 10.09 |
| 1600 | 1.01 | 8.73 | 3988.13 | 11.36 | 11.74 |
| 6400 | 1.18 | 33.04 | 8063.14 | 29.20 | 29.23 |
| 25600 | 1.26 | 133.75 | -- | 103.94 | 105.16 |
| 102400 | 1.43 | 531.67 | -- | 387.06 | 387.59 |
| 409600 | 2.00 | 2108.44 | -- | 1510.00 | 1509.49 |
| 1638400 | 4.95 | 8383.07 | -- | 5460.07 | 5283.01 |
| 6553600 | 10.89 | 34321.96 | -- | -- | -- |
| 11757483 | 18.20 | -- | -- | -- | -- |

Supplemental Table 5: Runtime results (in seconds) of different method at producing 1 synthetic individual datapoint over different sized increments of the human genome – from the 1000 genomes dataset. * GPU accelerated method

## Supplemental Section 8: Attribute Inference Experiment Diagram

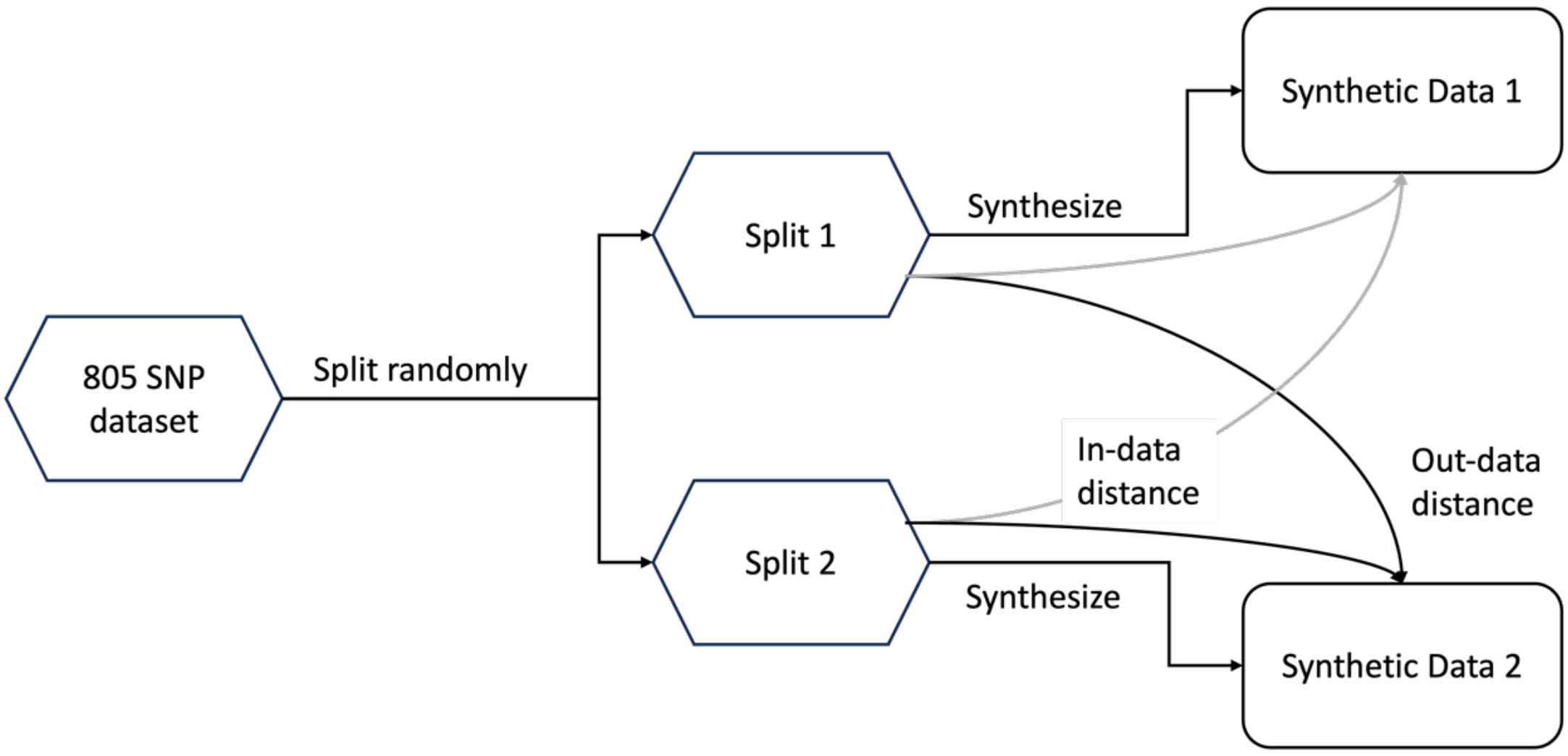

Supplemental Figure 9. Experimental process used in generating a data attribute inference attack. In this context the 805 SNP dataset was split into two equal subsets and synthetic data was generated from each, after this generation the median nearest neighbour hamming distance was calculated for each input to the output set that were involved in generating, and not, was calculated. This median distance from the input data to the synthetic data it was generated from is the 'in-data' distance, and the median distance from the input data to the synthetic data it was not generated from is the 'out-data' distance.

## Supplemental Section 9: Reproduction of Well Known Pharmacogenetic SNPs in Large Synthetic Datasets

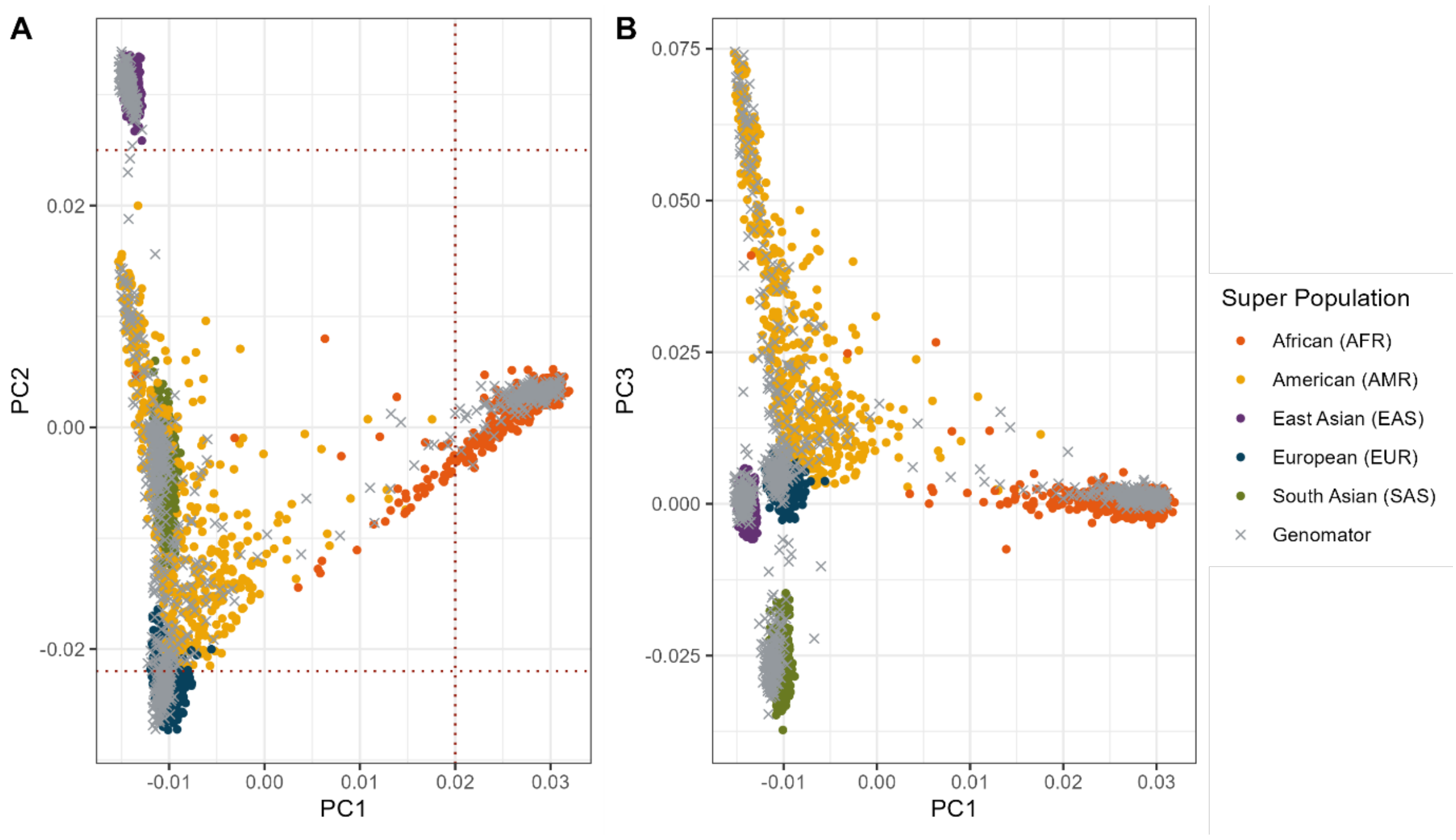

Supplemental Figure 10. Scatterplot of principal components (PC) 1 and 2 (A) and PCs 1 and 3 (B) from the Principal Component Analysis (PCA) of the 1000 Genomes Project (1KG) phase 3 (*n* = 3202) with Genomator generated samples (*n* = 1000) projected onto this space in black crosses. The 1KG samples are annotated according to the corresponding 'Super

Population' metadata. Thresholds to cluster the Genomator generated samples to the 1KG annotated 'Super Population' groups are highlighted in red.

| **Super Population** | **1KG (*n*)** | **1KG (%)** | **Genomator (*n*)** | **Genomator (%)** |
|---|---|---|---|---|
| **African (AFR)** | 893 | 27.89 | 250 | 25.0 |
| **East Asian (EAS)** | 585 | 18.27 | 180 | 18.0 |
| **European (EUR)** | 633 | 19.77 | 143 | 14.3 |
| **Other (SAS/AMR)** | 1091 | 34.07 | 427 | 42.7 |

Supplemental Table 6. Samples from the 1000 Genomes Project (1KG) (*n* = 3202) classified into four 'Super Population' based on their corresponding metadata. Genomator samples (*n* = 1000) were classified into the same four 'Super Population' based on their projection onto principal component analysis space (Supplemental Figure 10).

## Supplemental Section 10: Private vs Fictitious SNP quadruplets

| **Method** | **Private** | **Fictitious** |
|---|---|---|
| CRBM | 0.242 | 0.108 |
| | 0.308 | 0.146 |
| | 0.274 | 0.128 |
| | 0.221 | 0.104 |
| Markov | 0.402 | 0.175 |
| | 0.376 | 0.142 |
| | 0.325 | 0.095 |
| | 0.322 | 0.084 |
| | 0.352 | 0.120 |
| | 0.401 | 0.166 |
| | 0.347 | 0.108 |
| | 0.299 | 0.069 |
| | 0.310 | 0.075 |
| | 0.286 | 0.062 |
| RBM | 0.472 | 0.196 |
| | 0.446 | 0.175 |
| | 0.423 | 0.153 |
| | 0.436 | 0.157 |
| GAN | 0.148 | 0.052 |
| | 0.257 | 0.083 |
| | 0.203 | 0.070 |
| Genomator | 0.041 | 3.95E-04 |
| | 0.001 | 4.17E-05 |
| | 7.73E-05 | 3.34E-06 |
| | 1.29E-05 | 1.67E-06 |
| t-test(GENOMATOR, CRBM) | 1.18E-05 | 8.53E-06 |
| t-test(GENOMATOR, Markov) | 1.65E-09 | 9.21E-05 |
| t-test(GENOMATOR, RBM) | 4.85E-08 | 1.17E-06 |
| t-test(GENOMATOR, GAN) | 5.93E-04 | 1.29E-04 |
| Average CRBM | 0.261 | 0.121 |
| Average Markov | 0.342 | 0.109 |
| Average RBM | 0.444 | 0.170 |
| Average GAN | 0.202 | 0.068 |
| Average Genomator | 0.011 | 1.10E-04 |
| Min CRBM | 0.221 | 0.104 |
| Min Markov | 0.286 | 0.062 |
| Min RBM | 0.423 | 0.153 |
| Min GAN | 0.148 | 0.052 |
| Min Genomator | 1.29E-05 | 1.67E-06 |
| Max Genomator | 0.041 | 3.95E-04 |
| Percentage Improvement over CRBM | 96% | 100% |
| over Markov | 97% | 100% |
| over RBM | 98% | 100% |
| over GAN | 95% | 100% |
| Times improvement over CRBM | 5 | 262 |
| over Markov | 7 | 156 |
| over RBM | 10 | 388 |
| over GAN | 4 | 132 |

Supplemental Table 7: Likelihood of private vs fictitious quadruplets produced for a range of different parameters for each method, as well as the averages per method (where applicable). We note that the tools perform remarkably consistent, despite different parameters explored, i.e. Markov Chain (window sizes of 20,40,60,80,100,120,140,160,180,200), RBM (hidden layer sizes of 500,700 with learning rates of 0.005 and 0.01 to 1000 epochs) CRBM (hidden layer sizes of 500,600 with and without window overlaps of 300 and 400), GANs (trained to 20000 epochs with neural-layer-size multiplier 0.7,1.0 and 1.3), and Genomator (run with N=10->25 with Z=0->3 in unit increments).

## Supplemental Section 11: PCA of Genomator on Yelmen's larger datasets

The following **Supplemental Figures 11** and **12** show plotted PCA results produced by Genomator against the largest two principal components of the dataset, particularly the 10k and 65k datasets used in the study by Yelmen et al.[9].

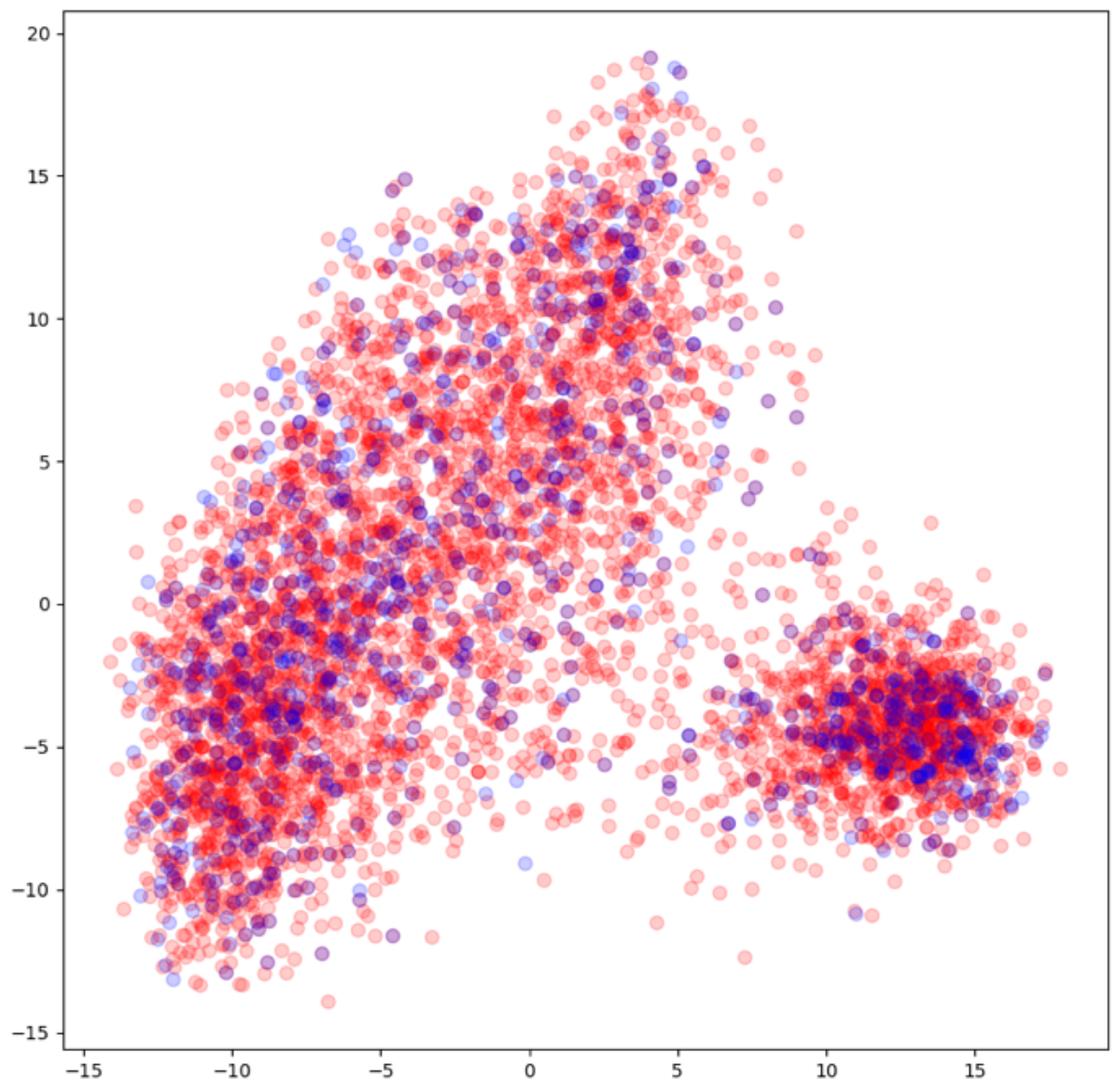

Supplemental Figure 11: PCA of genomator data (shown in blue) generated from the dataset of 10k haplotypes used in Yelmen et.al's 2023 paper, against the dataset itself (shown in red).

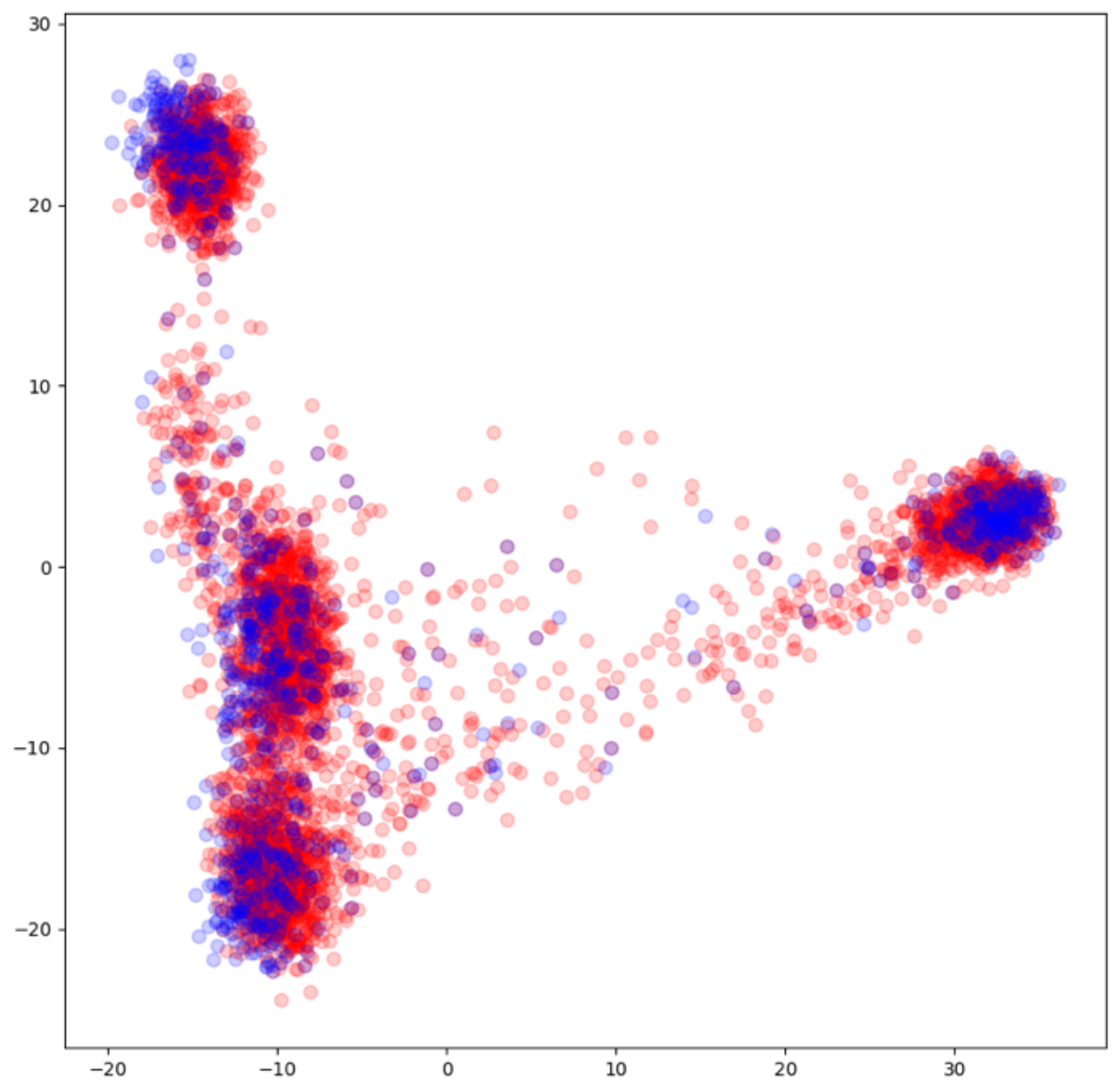

Supplemental Figure 12: PCA of genomator data (shown in blue) generated from the dataset of 65k haplotypes used in Yelmen et.al's 2023 paper, against the dataset itself (shown in red).

## Supplemental Section 12: Septuplets experiment

Like the Quadruplets experiment in the main paper, we consider extending the experiment to higher order, in this context we considered 'private' and 'fictitious' reproduction of septuplets of SNP combinations.

**Supplemental Figure** 13 shows likelihood of reproduction of different kinds of septuplets similar to that shown in the main paper of quadruplets.

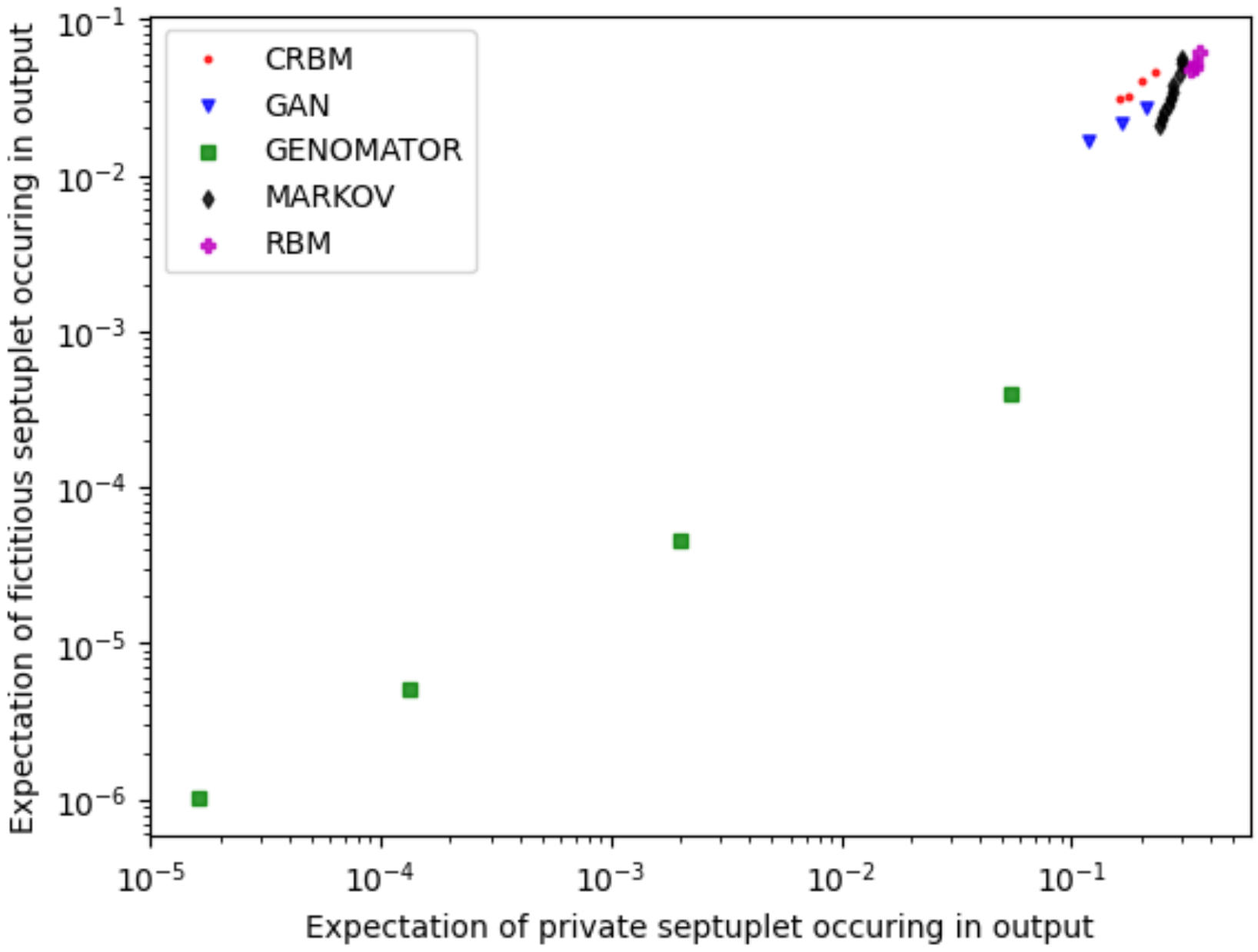

Supplemental Figure 13: plot showing the likelihood that the different methods will reproduce private and fictitious combinations (of order 7, ie septuplets) in the synthetic data, where a private combination is that that exists in exactly one input datum, and a fictitious combination is that which exists in no input datum. For different methods and a range of parameters detailed in the main text – specifically: Markov Chain (window sizes of 20,40,60,80,100,120,140,160,180,200), RBM (hidden layer sizes of 500,700 with learning rates of 0.005 and 0.01 to 1000 epochs) CRBM (hidden layer sizes of 500,600 with and without window overlaps of 300 and 400), GANs (trained to 20000 epochs with neural-layer-size multiplier 0.7,1.0 and 1.3), and Genomator(run with N=10->25 with Z=0->3 in unit increments).

## Supplemental Section 13: PCA for Z parameters

To visually show the effect of increasing Z on the accuracy attained by the generation method we show PCA obtained from Genomator with Z=0 and Z=6, in **Supplemental Figures 14** and **15** respectively. We note that N needed to be increased to support the possible generation using a higher Z value. The function of Z is to directly attenuate less common feature pairs and we can see that this creates the expected distortion by considering **Supplemental Figure 16**.

In **Supplemental Figure 16** a PCA is plotted of data that is produced by repeatedly taking clusters of input (of size 20) and producing data by taking the majority SNP at each location – this process simply attenuates less common SNPs which results in a similar distortion to that shown in **Supplemental Figure 15**.

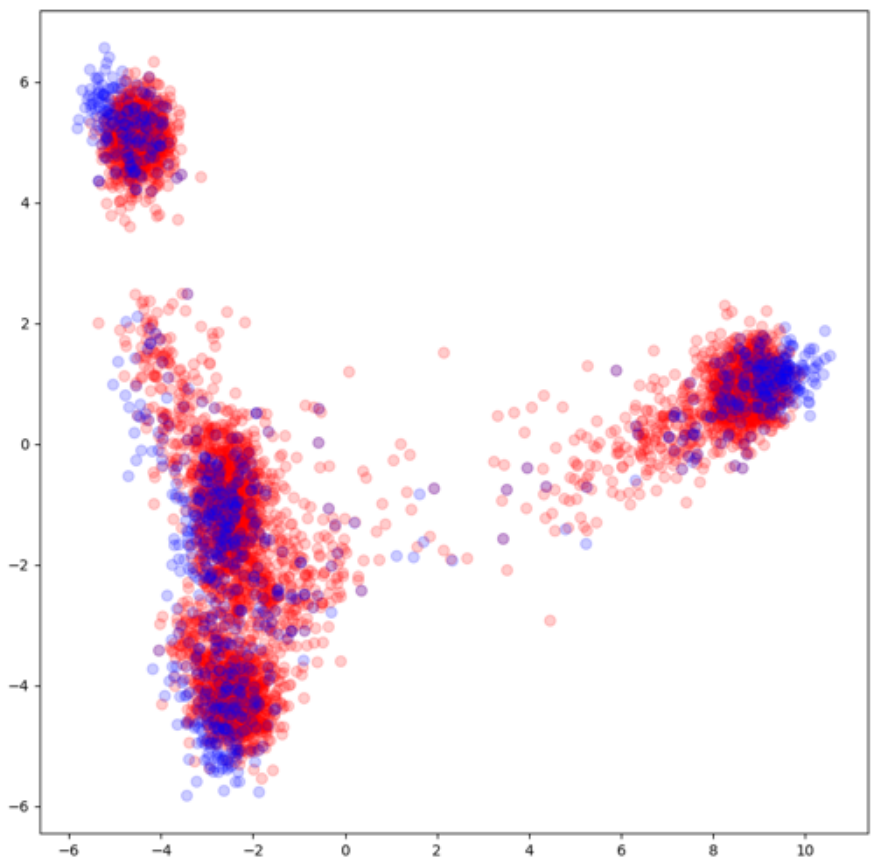

Supplemental Figure 14: PCA of Genomator output (blue) on principle components of Yelman's 805 haplotype data (red), set with N=10, Z=0

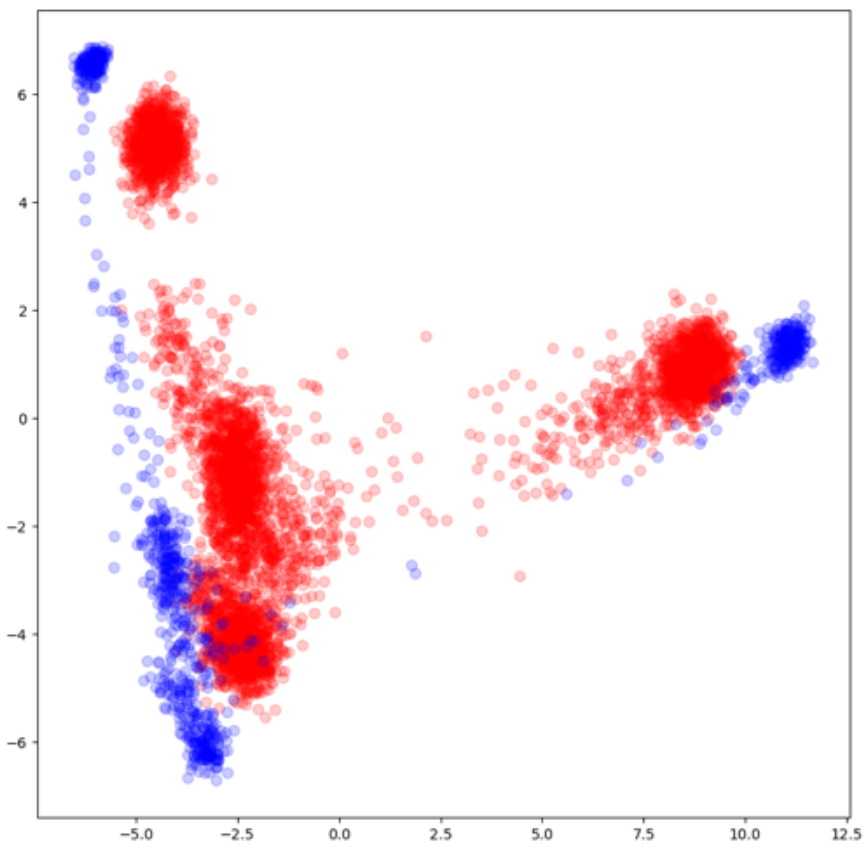

Supplemental Figure 15: PCA of Genomator output (blue) on principle components of Yelman's 805 haplotype data (red), set with N=45, Z=6

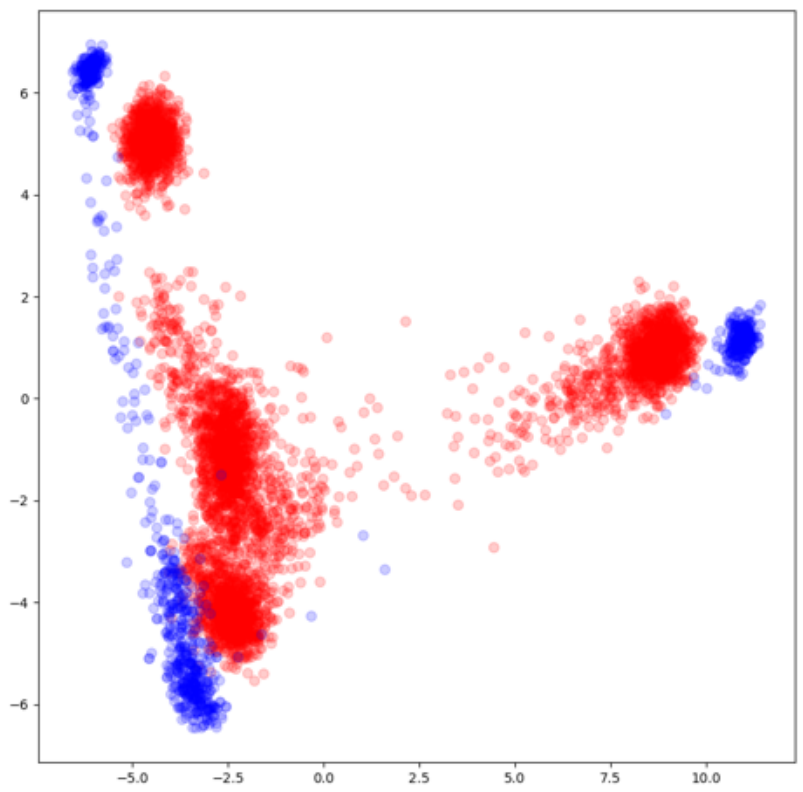

Supplemental Figure 16: PCA of genomes composed of the most common SNPs across input clusters of size 20, on the principle components of Yelman's 805 haplotype data

## Supplemental Section 14: Tabular Results for Attribute Inference Experiment

The Attribute Inference Experiment detailed in the main paper consisted of a range of different methods with different parameters, and for each, the median distance between the real data points and the nearest synthetic data point from the set generated from a dataset including them (the ‘in-data’ distance) and the median distance to the nearest synthetic data point from the set generated from a dataset not including them (the ‘out-data’ distance) difference, is reported in the following table.

| **METHOD** | **OPTION1** | **OPTION2** | **In-data distance** | **Out-data distance minus in-data distance** |
|---|---|---|---|---|
| **CRBM** | **Hidden-layer-size** | | | |
| CRBM | 100 | | 0.382608695652174 | 0 |
| CRBM | 200 | | 0.381366459627329 | -0.00124223602484469 |
| CRBM | 300 | | 0.368944099378882 | 0.000621118012422317 |
| CRBM | 400 | | 0.366459627329193 | 0 |
| CRBM | 500 | | 0.366459627329193 | 0.00124223602484469 |
| CRBM | 600 | | 0.363975155279503 | 0 |
| CRBM | 700 | | 0.362732919254658 | 0 |
| CRBM | 800 | | 0.363975155279503 | 0 |
| CRBM | 900 | | 0.363975155279503 | 0 |
| CRBM | 1000 | | 0.362732919254658 | 0.00124223602484469 |
| CRBM | 1100 | | 0.361490683229814 | 0.00124223602484475 |
| CRBM | 1200 | | 0.361490683229814 | 0 |
| CRBM | 1300 | | 0.363975155279503 | 0 |
| CRBM | 1400 | | 0.365217391304348 | 0 |
| CRBM | 1500 | | 0.362732919254658 | 0.00124223602484469 |
| CRBM | 1600 | | 0.362732919254658 | 0 |
| CRBM | 1700 | | 0.363975155279503 | 0.00124223602484475 |
| CRBM | 1800 | | 0.362732919254658 | 0 |
| CRBM | 1900 | | 0.363975155279503 | 0 |
| CRBM | 2000 | | 0.362732919254658 | 0 |
| **GAN** | **Neural layer size multiplier** | | | |
| GAN | 0.1 | | 0.270807453416149 | 0.00248447204968943 |
| GAN | 0.4 | | 0.273291925465838 | 0.00372670807453412 |
| GAN | 0.7 | | 0.272049689440994 | 0.00372670807453418 |
| GAN | 1 | | 0.277018633540373 | 0.00372670807453418 |
| GAN | 1.3 | | 0.274534161490683 | 0.00496894409937887 |
| GAN | 1.6 | | 0.272049689440994 | 0.00496894409937887 |
| GAN | 1.9 | | 0.277018633540373 | 0.00496894409937892 |
| GAN | 2.2 | | 0.277018633540373 | 0.00496894409937892 |
| GAN | 2.5 | | 0.277018633540373 | 0.00496894409937892 |
| GAN | 2.8 | | 0.270807453416149 | 0.00372670807453412 |
| GAN | 3.1 | | 0.26832298136646 | 0.00372670807453418 |
| GAN | 3.4 | | 0.273291925465838 | 0.00372670807453412 |
| GAN | 3.7 | | 0.263354037267081 | 0.00248447204968943 |
| GAN | 4 | | 0.259627329192547 | 0.00372670807453412 |
| GAN | 4.3 | | 0.26583850931677 | 0.00248447204968943 |
| GAN | 4.6 | | 0.26832298136646 | 0.00372670807453418 |
| **GENOMATOR** | **N** | **Z** | | |
| GENOMATOR | 15 | 1 | 0.204968944099379 | 0.0347826086956522 |
| GENOMATOR | 20 | 2 | 0.212422360248447 | 0.0248447204968944 |
| GENOMATOR | 25 | 3 | 0.216149068322981 | 0.0198757763975155 |
| GENOMATOR | 30 | 4 | 0.217391304347826 | 0.0186335403726708 |
| GENOMATOR | 35 | 5 | 0.219875776397516 | 0.0161490683229814 |
| GENOMATOR | 40 | 6 | 0.22111801242236 | 0.0149068322981366 |
| GENOMATOR | 45 | 7 | 0.222360248447205 | 0.0136645962732919 |
| GENOMATOR | 50 | 8 | 0.222360248447205 | 0.0124223602484472 |
| GENOMATOR | 60 | 10 | 0.224844720496894 | 0.0111801242236025 |
| GENOMATOR | 70 | 12 | 0.226086956521739 | 0.00993788819875777 |
| GENOMATOR | 80 | 14 | 0.227329192546584 | 0.00745341614906833 |
| GENOMATOR | 90 | 16 | 0.228571428571429 | 0.00745341614906833 |
| GENOMATOR | 100 | 18 | 0.228571428571429 | 0.00745341614906833 |
| GENOMATOR | 120 | 22 | 0.229813664596273 | 0.00621118012422361 |
| GENOMATOR | 140 | 26 | 0.231055900621118 | 0.0049689440993789 |
| GENOMATOR | 160 | 30 | 0.232298136645963 | 0.00372670807453415 |
| GENOMATOR | 180 | 34 | 0.233540372670807 | 0.00372670807453415 |
| GENOMATOR | 200 | 38 | 0.233540372670807 | 0.00372670807453415 |
| GENOMATOR | 240 | 46 | 0.233540372670807 | 0.00372670807453415 |
| GENOMATOR | 280 | 54 | 0.236024844720497 | 0.00248447204968943 |
| **MARKOV** | **Window size** | | | |
| MARKOV | 2 | | 0.35527950310559 | 0.00124223602484469 |
| MARKOV | 4 | | 0.351552795031056 | 0 |
| MARKOV | 6 | | 0.349068322981366 | 0 |
| MARKOV | 8 | | 0.342857142857143 | 0 |
| MARKOV | 10 | | 0.339130434782609 | 0 |
| MARKOV | 12 | | 0.329192546583851 | 0.00124223602484469 |
| MARKOV | 14 | | 0.319254658385093 | 0 |
| MARKOV | 16 | | 0.309316770186335 | 0.00248447204968943 |
| MARKOV | 18 | | 0.296894409937888 | 0.00745341614906836 |
| MARKOV | 20 | | 0.277018633540373 | 0.0211180124223603 |
| MARKOV | 22 | | 0.252173913043478 | 0.0422360248447205 |
| **RBM** | **Hidden layer size** | **Learning rate** | | |
| RBM | 100 | 0.005 | 0.28944099378882 | 0.00124223602484475 |
| RBM | 100 | 0.01 | 0.288198757763975 | 0.00124223602484475 |
| RBM | 200 | 0.005 | 0.288198757763975 | 0 |
| RBM | 200 | 0.01 | 0.288198757763975 | 0.00124223602484475 |
| RBM | 300 | 0.005 | 0.288198757763975 | 0 |
| RBM | 300 | 0.01 | 0.28695652173913 | 0 |
| RBM | 400 | 0.005 | 0.288198757763975 | 0 |
| RBM | 400 | 0.01 | 0.285714285714286 | 0.00248447204968943 |
| RBM | 500 | 0.005 | 0.288198757763975 | 0.00124223602484475 |
| RBM | 500 | 0.01 | 0.285714285714286 | 0.00124223602484475 |
| RBM | 600 | 0.005 | 0.28695652173913 | 0.00124223602484469 |
| RBM | 600 | 0.01 | 0.285714285714286 | 0.00124223602484475 |
| RBM | 700 | 0.005 | 0.28695652173913 | 0.00124223602484469 |
| RBM | 700 | 0.01 | 0.285714285714286 | 0.00124223602484475 |
| RBM | 800 | 0.005 | 0.288198757763975 | 0 |
| RBM | 800 | 0.01 | 0.284472049689441 | 0.00248447204968943 |
| RBM | 900 | 0.005 | 0.28695652173913 | 0.00124223602484469 |
| RBM | 900 | 0.01 | 0.284472049689441 | 0.00248447204968943 |
| RBM | 1000 | 0.005 | 0.28695652173913 | 0.00124223602484469 |
| RBM | 1000 | 0.01 | 0.285714285714286 | 0.00124223602484475 |
| RBM | 1100 | 0.005 | 0.28695652173913 | 0.00124223602484469 |
| RBM | 1100 | 0.01 | 0.284472049689441 | 0.00248447204968943 |
| RBM | 1200 | 0.005 | 0.28695652173913 | 0 |
| RBM | 1200 | 0.01 | 0.284472049689441 | 0.00248447204968943 |
| RBM | 1300 | 0.005 | 0.28695652173913 | 0.00124223602484469 |
| RBM | 1300 | 0.01 | 0.284472049689441 | 0.00248447204968943 |
| RBM | 1400 | 0.005 | 0.288198757763975 | 0 |
| RBM | 1400 | 0.01 | 0.284472049689441 | 0.00248447204968943 |
| RBM | 1500 | 0.005 | 0.28695652173913 | 0.00124223602484469 |
| RBM | 1500 | 0.01 | 0.284472049689441 | 0.00248447204968943 |
| RBM | 1600 | 0.005 | 0.28695652173913 | 0.00124223602484469 |
| RBM | 1600 | 0.01 | 0.284472049689441 | 0.00248447204968943 |
| RBM | 1700 | 0.005 | 0.28695652173913 | 0 |
| RBM | 1700 | 0.01 | 0.285714285714286 | 0.00248447204968943 |
| RBM | 1800 | 0.005 | 0.28695652173913 | 0.00124223602484469 |
| RBM | 1800 | 0.01 | 0.284472049689441 | 0.00248447204968943 |
| RBM | 1900 | 0.005 | 0.28695652173913 | 0.00124223602484469 |
| RBM | 1900 | 0.01 | 0.284472049689441 | 0.00248447204968943 |
| RBM | 2000 | 0.005 | 0.28695652173913 | 0.00124223602484469 |
| RBM | 2000 | 0.01 | 0.284472049689441 | 0.00248447204968943 |

Supplemental Table 8: The performance of each of the methods and their parameters on the Attribute Inference Experiment

The minimum Euclidean distance on Figure 4, for each of the methods to the ideal point (0,0) was evaluated and shown in the following table:

| Method | Minimum Distance to ideal | Percent Distance increase over Genomator |
|---|---|---|
| CRBM | 0.361490683229814 | 73.87% |
| MARKOV | 0.25965407452364 | 24.89% |
| GAN | 0.25568645684577 | 22.98% |
| RBM | 0.284482898705489 | 36.83% |
| GENOMATOR | 0.207899249428393 | 0.0% |

Supplemental Table 9: The minimum Euclidean distance to the ideal point at the origin of Figure 4 – where the y-axis is difference between in-data and out-data score, and x-axis is in-data score. Shown also is the proportion reduction in this minimum distance which Genomator achieves calculated from these minimal distance scores.